\documentclass[sigconf]{acmart}

\usepackage{enumitem}
\usepackage{stfloats}

\AtBeginDocument{%
  \providecommand\BibTeX{{%
    \normalfont B\kern-0.5em{\scshape i\kern-0.25em b}\kern-0.8em\TeX}}}

\copyrightyear{2023}
\acmYear{2023}
\setcopyright{acmlicensed}\acmConference[MM '23]{Proceedings of the 31st ACM International Conference on Multimedia}{October 29-November 3, 2023}{Ottawa, ON, Canada}
\acmBooktitle{Proceedings of the 31st ACM International Conference on Multimedia (MM '23), October 29-November 3, 2023, Ottawa, ON, Canada}
\acmPrice{15.00}
\acmDOI{10.1145/3581783.3611957}
\acmISBN{979-8-4007-0108-5/23/10}

\acmSubmissionID{1177}

\graphicspath{ {./fig/} }
\begin{document}

%\title{ABF: Alpha Blending Field for Editing with Automatic Color Palette}
%\title{RecolorNeRF: Jointly-optimized Layer Decomposition for \\ Efficient Color Editing of Neural Radiance Fields}
%\title{RecolorNeRF: Efficient Color Editing of Layer Decomposed Radiance Fields}
\title[RecolorNeRF: Layer Decomposed Radiance Fields for Efficient Color Editing of 3D Scenes]{RecolorNeRF: Layer Decomposed Radiance Fields \\for Efficient Color Editing of 3D Scenes}

\author{Bingchen Gong}
\authornote{Both authors contributed equally. Corresponding email to: gongbingchen@gmail.com}
\email{gongbingchen@gmail.com}
\orcid{0000-0001-6459-6972}
\author{Yuehao Wang}
\authornotemark[1]
\email{yhwang@link.cuhk.edu.hk}
\orcid{0009-0003-3144-128X}
\affiliation{%
  \institution{The Chinese University of Hong Kong}
%  \streetaddress{P.O. Box 1212}
%  \city{Hong Kong}
%  \country{Hong Kong}
}

\author{Xiaoguang Han}
\email{hanxiaoguang@cuhk.edu.cn}
\orcid{0000-0003-0162-3296}
\affiliation{%
	\institution{The Chinese University of Hong Kong (Shenzhen)}
%  \streetaddress{P.O. Box 1212}
%  \city{Shenzhen}
%  \country{China}
}

\author{Qi Dou}
\email{qidou@cuhk.edu.hk}
\orcid{0000-0002-3416-9950}
\affiliation{%
	\institution{The Chinese University of Hong Kong}
%  \streetaddress{P.O. Box 1212}
%  \city{Hong Kong}
%  \country{Hong Kong}
}


%%%%%%%%% ABSTRACT
\begin{abstract}
   Radiance fields have gradually become a main representation of media. Although its appearance editing has been studied, how to achieve view-consistent recoloring in an efficient manner is still under explored. We present RecolorNeRF, a novel user-friendly color editing approach for the neural radiance fields. Our key idea is to decompose the scene into a set of pure-colored layers, forming a palette. By this means, color manipulation can be conducted by altering the color components of the palette directly. To support efficient palette-based editing, the color of each layer needs to be as representative as possible. In the end, the problem is formulated as an optimization problem, where the layers and their blending weights are jointly optimized with the NeRF itself. Extensive experiments show that our jointly-optimized layer decomposition can be used against multiple backbones and produce photo-realistic recolored novel-view renderings. We demonstrate that RecolorNeRF outperforms baseline methods both quantitatively and qualitatively for color editing even in complex real-world scenes.
   Our code and more results are available at \url{https://sites.google.com/view/recolornerf}.
  % The ability to recolor a scene in neural implicit representation is vitally important for film production and most wanted by colorists, however, is still yet explored. We present RecolorNeRF, a novel palette-based color editing method of the neural radiance field with jointly optimized layer decomposition. In RecolorNeRF, we decompose the scene into multiple pure-colored layers where each layer has a corresponding learnable palette. Multiple layers are stacked together over the alpha composition by layer opacity to produce the radiance color. The layer decomposition is view-dependent and Multi-view consistency is guaranteed by NeRF rendering. To capture the most representative colors of the scene into the palette, we design a novel convex-hull regularization to guide the palette converging to existing colors in the scene and jointly optimize the palette with layer opacity. Sparsity constraints are applied to opaque values to achieve optimal decomposition. Our method allows users to recolor NeRF on-the-fly by just altering the colors in the palette. Extensive experiments show that our jointly-optimized layer decomposition can be used against multiple backbones and produce photo-realistic recolored novel-view renderings. We demonstrate that RecolorNeRF outperforms baseline methods both quantitatively and qualitatively for color editing even in complex real-world scenes.

\end{abstract}

\begin{CCSXML}
  <ccs2012>
  <concept>
      <concept_id>10010147.10010178.10010224.10010245.10010254</concept_id>
      <concept_desc>Computing methodologies~Reconstruction</concept_desc>
      <concept_significance>500</concept_significance>
  </concept>
  <concept>
      <concept_id>10010147.10010371.10010382.10010385</concept_id>
      <concept_desc>Computing methodologies~Image-based rendering</concept_desc>
      <concept_significance>300</concept_significance>
  </concept>
  <concept>
      <concept_id>10010147.10010371.10010396.10010401</concept_id>
      <concept_desc>Computing methodologies~Volumetric models</concept_desc>
      <concept_significance>300</concept_significance>
  </concept>
  </ccs2012>
\end{CCSXML}

\ccsdesc[500]{Computing methodologies~Reconstruction}
\ccsdesc[300]{Computing methodologies~Image-based rendering}
\ccsdesc[300]{Computing methodologies~Volumetric models}

\keywords{radiance fields, recoloring, visual editing, neural rendering, palettes}


\begin{teaserfigure}
  \centering
  \includegraphics[trim={5 5 0 10},clip,width=\textwidth]{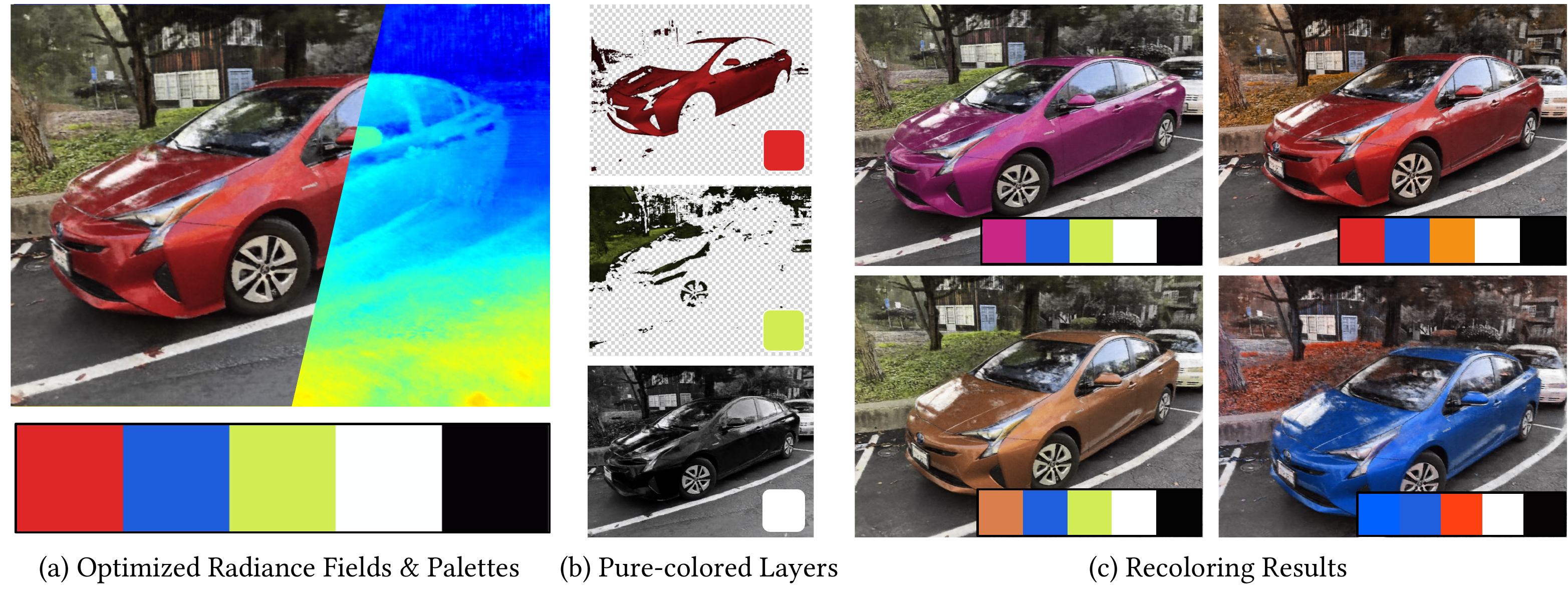}
  \caption{Our method offers a user-friendly approach for recoloring 3D scenes. The scene is represented as a radiance field that is layer-decomposed and jointly optimized with an associated palette. Each color in the palette corresponds to a pure-colored layer. This design enables users to alter the color of each layer independently, producing high-fidelity recoloring results.}
  \label{fig:teaser}
  \vspace*{-2px}
\end{teaserfigure}

\maketitle

%%%%%%%%% BODY TEXT
\section{Introduction}
\label{sec:intro}

% \begin{figure}[t]
%   \centering
%   \includegraphics[trim={0 0 0 0},clip,width=\linewidth]{fig/23_ARXIV_TEASER.pdf}
%   \caption{Our method recolors an image by automatically decomposing it into layers and recoloring each layer independently. The user can specify the color of each layer, and our method automatically blends the layers.}
%   \vspace*{-10px}
% \end{figure}

\begin{figure*}[t]
	\centering
	\includegraphics[trim={0 5 0 5},clip,width=\textwidth]{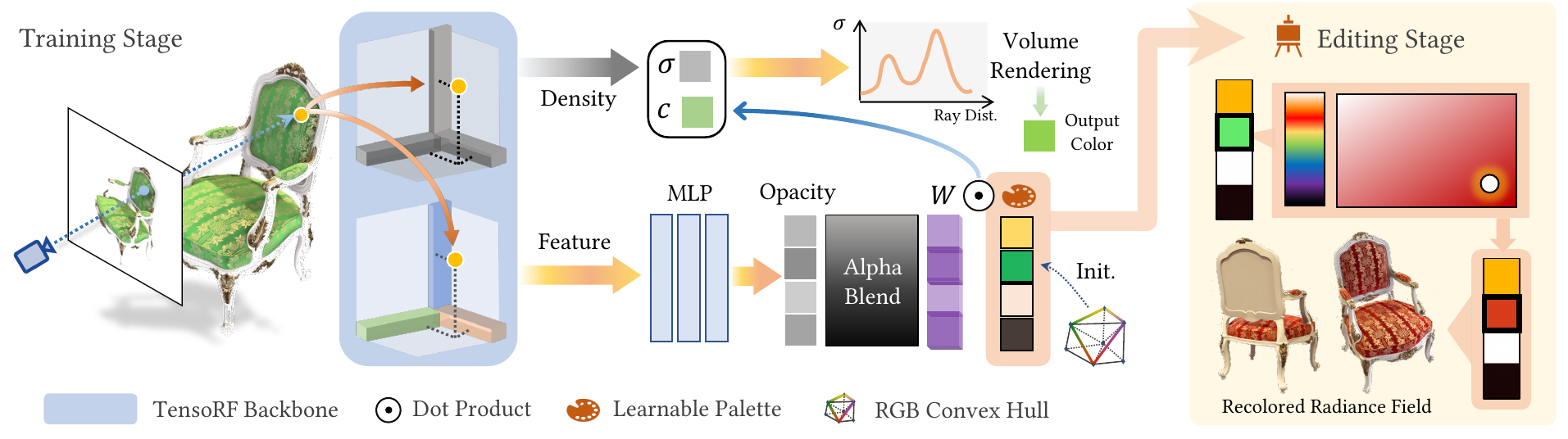}
	\vspace*{-18px}
	\caption{Overview of our method.
		During the training stage, the TensoRF \cite{chen2022tensorf} backbone and the learnable palette will be optimized to represent the color of any query point through alpha compositing over the learned palette.
		In the editing stage, users can simply replace a color in the palette with a new one to recolor the entire radiance field.}
	% We first decompose the NeRF model into layers with the color palette. Then we joint-optimize the layers with the palette to fit the layer decomposed radiance field on a specific scene. Finally, we edit the palette and render the recolored NeRF model to generate photo-realistic images. The decomposed layer are stacked together on the user-specified order with alpha blending to support efficient editing on complete scene.
	\label{fig:network}
	\vspace*{-12px}
\end{figure*}

Neural radiance fields (NeRF) have been proven to be a powerful representation of 3D scenes, which likely become a next-generation media form in the future, analogous to images and videos. To this end, supporting editing in such a new representation is critical.
Very recently, several works on this topic have emerged, exploring editable NeRF in respect of scene composition \cite{zhang2021editable,tang2022compressible}, geometry deformation \cite{liu2021editing,yuan2022nerf,munkberg2022extracting,xu2022deforming}, appearance editing \cite{yang2021learning,huang2022hdr,mildenhall2022nerf}, style transfer \cite{sun2022nerfeditor,chen2022upst,fan2022unified,zhang2022arf} etc.
%Recoloring, as one type of appearance editing, usually indicates resetting some specific color tones of a scene for enhancement or correction, which plays an important role in film production and artistic creation. As an example exhibited in Fig.~\ref{fig:teaser}, by recoloring, the red car can be repainted into a blue one with photo-realism maintained.
Recoloring is a sort of appearance editing that typically entails resetting a scene's precise color tones for improvement or correction. It is a crucial process in the creation of both artistic and cinematic works. For example, using recoloring, the red car in Fig.~\ref{fig:teaser} may be colored blue while maintaining photorealism.

% Among all existing approaches to recoloring an image, palette-based color editing (PCE) \cite{aksoy2017unmixing, zhang2017palette, wang2019improved} provides the most intuitive means for efficient user interaction. Specifically, PCE extracts a set of representative colors from the image, namely a palette. For each item of the palette, we define a corresponding image layer with a constant color value of the item. The main task of this step is to determine the blending weights of these layers to reconstruct the given image. When editing color, we simply alter the color of each layer and can precisely control the color component of target objects, e.g., tuning green pigments to recolor grass and leaves in an image of ``forest''.

Among all existing approaches to recoloring an image, palette-based color editing (PCE) \cite{aksoy2017unmixing, zhang2017palette, wang2019improved} stands out due to its most intuitive and efficient user interaction.
%PCE will first extract a set of representative colors from the image to form a palette. Each item of the palette is associated with a pure-colored layer in the corresponding color. The image will be decomposed as a blending of the layers by solving the weight of each palette item. After that, we can simply alter the palette to precisely edit colors in the image, e.g., tuning green layers to recolor grass and leaves in a ``forest'' image.
PCE first extracts a palette of representative colors from the image. Each color corresponds to a pure-colored layer. The image will be decomposed as a blending of the layers by solving the weight of each palette item. After that, we can simply alter the palette to precisely edit the image's colors, such as tuning green layers to recolor grass and leaves in a ``forest'' image.

%As one of the SOTA methods of PCE on 2D images, Tan et al. \cite{tan2015decomposing,tan2016decomposing,tan2018efficient} proposed a convex-hull simplification strategy for palette extraction. With the obto deained palette, layer decomposition is then formulated as an optimization problem. To make the problem solvable, the sparsity of the blending weights is assumed.

In this paper, we propose a novel method, RecolorNeRF, to conduct photo-realistic PCE on NeRF representations, which to our knowledge is the first attempt using a learnable palette for layer-wise decomposition of 3D scenes.
%\footnote{The concurrent works PaletteNeRF (https://palettenerf.github.io/ and https://arxiv.org/abs/2212.12871) are not taken into account.}.
Although \cite{tojo2022recolorable} has performed NeRF recoloring based on palettes, its rendering results are compromised by its lossy posterization.
% can only enable color editing after posterization, making the results unrealistic.
% As known, the NeRF of a scene is commonly reconstructed from multi-view images. Thus
Another possible approach to adapting PCE to NeRF is to first extract palettes from the pixels of all input images, using the heuristic method in \cite{tan2016decomposing}, and then separately conduct layer decomposition and color editing for each novel view rendered from a pre-trained NeRF model. Despite being straightforward to implement, this solution suffers from exhausted per-view decomposition, poor view inconsistency, and 3D-agnostic palette extraction.
% 3 major issues: 1) this means of recoloring compromises into a post-process of NeRF rendering, which causes expensive computational costs. 2) as each view is independently processed, the results tend to lack view consistency. Third, the palette extraction is achieved by a heuristic method, which may make the palette color less representative and the layer decomposition not clean enough therefore interfering with the color manipulation.
%This may cause the palette color to be not representative and the layer decomposition not clean enough, thus interfering with the color manipulation.

To address the aforementioned issues, our key idea is to optimize the palette, the layer decomposition, and the volumetric radiance fields in a unified framework. To deal with complicated scenes, we follow \cite{richardt2014vectorising,tan2016decomposing} to use ``over'' composition as the imaging formulation. Specifically, the appearance of each point in a 3D scene is represented by an alpha blending of a set of ordered pure-colored layers, which form a palette for editing. To achieve this, we model the layer opacity as a volumetric alpha field for each layer.
% For each pigment layer, we define a volumetric alpha field modeled in an MLP, similar to the ordinary radiance field.
% Note that, different layers use different MLPs. In this sense, we only need to optimize those MLPs for the blending weights which can also be jointly optimized with the MLPs for the density field. As known, all previous PCE methods conducted palette extraction independently.
Our proposed approach also includes the first trial of geometric palette optimization, which  regularizes the palette to convexly span the color space of the 3D scene. Furthermore, in order to encourage the independence and representativeness of the learnable palette, sparsity of the blending weights is imposed through our novel soft sparsity norm and order-dependent weighting scheme.

% To ease the joint optimization problem, two new designs are proposed: 1) To enable a small number of palette colors to accurately represent the whole scene, a novel convex-hull regularization is presented. 2) In order to make the palette color more representative, sparsities on the blending weights are used as usual. To further increase the ability to model complex scenes, a novel order-aware weighting scheme is presented for the sparsity constraint.

Our RecolorNeRF can robustly decompose 3D scenes into multiple layers and enable diverse high-fidelity recoloring of complex 3D scenes in an efficient manner, without any restriction to backbone NeRF models, fine-tuning, and additional deep feature extraction. Our comparison experiments and a large-group user study also validate the superiority of our approach over other color editing methods for 3D scenes.
In summary, our main contributions include:
\begin{itemize}[leftmargin=*]
  \item To our knowledge, we are the first attempt to perform photo-realistic palette-based color editing in NeRF representations via layer decomposition.
  %We are the first learnable palette-based editing method to recolor any neural radiance fields by decomposition and optimize the implicit representation with a 3D field that learns the blending of layers from the input images.
  \item We propose novel convex hull regularizations to jointly optimize the palette and the scene representation.
  % for which a novel convex-hull regulation is designed to make it solvable.
  %We decompose the scene into layers with pure color and jointly optimize the opaque value with the color palette. The layers are stacked together with alpha blending, enabling the recoloring of the NeRF model on-the-fly by altering the layer's palette.
  % \item The whole RecolorNeRF framework is carefully designed, allowing the color schemes can be efficiently edited even for complicated scenes.
  \item Extensive recoloring experiments show that our carefully designed approach supports various complex scenes and different NeRF backbones.
  %Extensive experiments show that RecolorNeRF can decompose the implicit representation to produce semantic reasonable layers and recolor the NeRF model with photo-realistic quality while preserving multiview consistency.
\end{itemize}

% \begin{figure*}[t]
%   \centering
%   \includegraphics[trim={0 10 0 10},clip,width=0.95\textwidth]{23_SIG_FRAMEWORK.pdf}
%   \caption{Overview of our method. We first decompose the NeRF model into layers with the color palette. Then we joint-optimize the layers with the palette to fit the layer decomposed radiance field on a specific scene. Finally, we edit the palette and render the recolored NeRF model to generate photo-realistic images. The decomposed layer are stacked together on the user-specified order with alpha blending to support efficient editing on complete scene.}
%   \label{fig:overview}
%   \vspace*{-10px}
% \end{figure*}

%------------------------------------------------------------------------
\section{Related Work}
\label{sec:related}

\begin{figure*}[t]
	\centering
	\includegraphics[trim={0 12 0 20},clip,width=\textwidth]{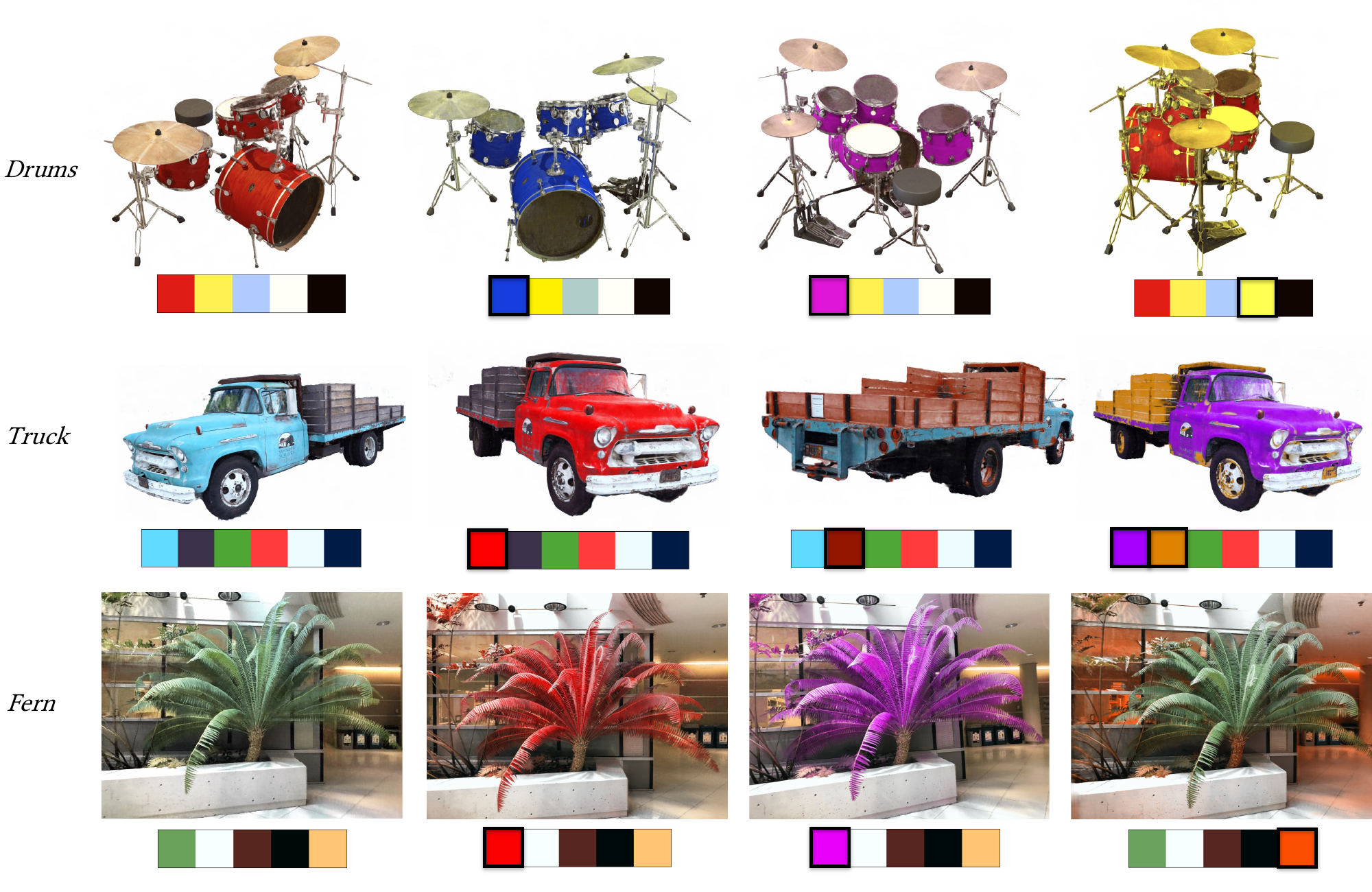}
	\vspace*{-20px}
	\caption{Gallery of our color editing results. The first image of each row is the reference before editing along with the beneath optimized palette. The other 3 images showcase 3 examples of color editing with the corresponding edited palettes.}
	\label{fig:res_vis1}
	\vspace*{-12px}
\end{figure*}

% In this section, we review related work to provide context for our method. We first overview the NeRF editing model variants. Then we review the existing image color editing methods. Finally, we review the methods that extract the palette and decompose the image.

%-------------------------------------------------------------------------
\subsection{NeRF Editing}

%Latent [NeRFEditor]
%Style [UPST-NeRF] [ARF]

Editing NeRFs is challenging due to their implicit scene representations. Existing works \cite{yang2021learning, liu2021editing, zhuang2022mofanerf, zheng2022editablenerf, xu2022signal}  support a few types of purely implicit editing on the target objects.
NeRF-editing \cite{yuan2022nerf} extracts explicit meshes as a manipulable proxy and transfers the proxy movements back into the implicit representations.
Approaches of \cite{zhang2021editable,tang2022compressible} present solutions to entity placement and scene composition in NeRF-represented scenes.
NVDiffRec \cite{munkberg2022extracting} extracts editable BRDF materials by disentangling 3D geometry and lighting effects from multi-view captures.
EditNeRF \cite{liu2021editing} and CodeNeRF \cite{yang2021learning} are the recent NeRF models that encode simple objects' shape and appearance into interpolative latent code.
HDR-NeRF \cite{huang2022hdr} and RawNeRF \cite{mildenhall2022nerf} use linear raw images as input rather than the post-processed ones, allowing controllable tone mapping during view synthesis.
CLIP-NeRF \cite{wang2022clip} and DFFs \cite{kobayashi2022distilledfeaturefields} support text prompt or exemplar image-guided editing by leveraging the joint vision-language embedding of the CLIP model \cite{radford2021learning}.
UPST-NeRF \cite{chen2022upst}, INS \cite{fan2022unified} and ARF \cite{zhang2022arf} propose different style transferring schemes and stylize NeRF models into reference images.
NeRFEditor \cite{sun2022nerfeditor} encodes novel-view images into the hidden space of StyleGAN and enables interactive style editing on NeRF models.

Most of the above methods supporting style and appearance editing adopt additional encoders to extract high-level appearance code, while none of them decomposes the implicit representations in color space and endows intrinsic physical meanings to the decomposition.
In contrast, our method can facilitate photo-realistic editing on both complex scenes and single objects without extra semantic-level features.
The editing mode of our approach is more controllable and flexible, featuring precise color tuning and user-designed palettes.
During the preparation of this paper, we noticed that there is a concurrent work PaletteNeRF \cite{kuang2022palettenerf} which also supports palette-based editing of NeRF models.
%They use a precomputed palette from existing works and learn direct weights on the palette.
Our method is different from theirs mainly in that we tend to decompose scenes into sparser layers using alpha blending, jointly optimize palettes and scene representations via more concise modeling, and allow the users to control the composition order of layers against a tailored palette. We include a result comparison between our method and PaletteNeRF in our supplementary materials to highlight the differences.

%-------------------------------------------------------------------------
\subsection{Color Editing}

Color editing on images is a long-standing problem and has been well-studied in many scenarios. The challenge of color editing mainly lies in the extraction of editable color patterns. Failure of sorting out color patterns will cause the so-called ``color pollution'' issue, which implies color edits upon different targets are mixed up and produces contamination-like artifacts.
% In general, we can classify these scenarios as 1) transferring color schemes from reference images, 2) directly editing the color in a local region, or 3) directly editing the color schemes of an image.
%
Many approaches exploit user guidance in pattern extraction by drawing colored scribbles or coarse spatial markers. Such approaches aim to recolor the meaningful patterns in the image by propagating the desired color through the user's brushwork. Various algorithms \cite{levin2004colorization, lischinski2006interactive, pellacini2007appwand, an2008appprop, xu2009efficient, chen2012manifold} are designed to propagate the user-specified edits. The problem is often formulated as an energy minimization problem, where Euclidean distance or diffusion distance in feature space is adopted to measure the affinity among pixels.
Drawing scribbles demands higher costs on the user side than reference-based or palette-based methods since users need to provide meaningful guidance both in spatial and color domains. Color transfer is another form of reference-based color editing, in which the source image absorbs the color patterns of another reference image.
Early color transfer methods \cite{pitie2005n, reinhard2001color} change the colors of the source image based on the color distribution of a reference image. \citet{reinhard2001color} performs color transfer by matching the means and standard deviations of the source color distribution to the reference one. \citet{pitie2005n} explore a color mapping between source and reference images via color probability density function transferring. Histograms are wildly used to represent the color distributions of images.
% The color transfer is performed by matching the histograms of the source and reference images.
\citet{delon2005automatic} propose a color transfer method that segments the color histogram to construct palettes and warp the source color to the target by referring to the palettes.
% While powerful, these color transfer methods provide very few editing controls to users other than the choice of a reference image.

\subsection{Palette-based Recoloring}

Palette-based recoloring methods allow users to tune color components by modifying a finite set of colors, which is a trade-off solution between user control and recoloring capability.
A palette is a succinct representation that can describe the color scheme of an image.
The most essential step in palette-based recoloring is palette extraction.
Some previous methods fit predictive models of human palette preference and generate perceptual palettes from input images \cite{o2011color, lin2013modeling, cao2017mining, feng2018finding}.
A more straightforward approach to extract palettes for color editing is using the k-means algorithm to cluster the image colors in RGB space \cite{chang2015palette, zhang2017palette}. This captures the most prominent colors in the image.
Another palette-based approach consists of computing and simplifying the convex hull that
encloses all the color samples \cite{tan2016decomposing}. This yields more ``primary'' palettes that better represent the color gamut of the image.
% Inspired by the geometry structure of the color convex hull, we use the convex hull to constrain the color palette optimization in our RecolorNeRF.
%
Once we have the palette extracted, we can use it to represent the colors in the image by decomposing the image into layers, where each layer corresponds to an importance map of a palette color. The layer decomposition performs as a reverse procedure of layer composition.
Several previous methods are proposed to composite the layers simply via weighted additive mixing \cite{aksoy2017unmixing, zhang2017palette, tan2018efficient, wang2019improved}.
% Though it is simple, the additive mixing is not stacking the layers in order, like the Painters' algorithm.
One most common alpha blending operator ``over'' composite is also used in recent methods to composite order-dependent layers \cite{richardt2014vectorising,tan2015decomposing,tan2016decomposing}. Our method uses alpha blending, allows users control of palette order, and facilitates sparser layer decomposition.

%------------------------------------------------------------------------
\section{Preliminaries}

% This section introduces preliminaries of the NeRF model and the color palette-based decomposition, defining the formulation of implicit representation and alpha composition used in RecolorNeRF. Fig.\ref{fig:overview} shows the overview of RecolorNeRF. We first decompose the NeRF model into layers with the color palette. Then we joint-optimize the layers to recolor the NeRF model. Finally, we render the recolored NeRF model to generate photo-realistic images.

\subsection{Neural Radiance Fields}

Neural Radiance Fields are introduced by \cite{mildenhall2021nerf} as a differentiable rendering model that can be trained to synthesize photo-realistic novel views by taking multi-view images as the input. Ordinary NeRFs can be regarded as an MLP function that maps coordinates into color and density values in the form of $F_{\Theta}(\mathbf{x}, \mathbf{d}) \rightarrow \left (\mathbf{c}, \sigma\right)$,
%
% \begin{equation}
% \label{eq:nerf_func}
% F_{\Theta}(\mathbf{x}, \mathbf{d}) \rightarrow \left (\mathbf{c}, \sigma\right)
% \end{equation}
%
where $\Theta$ represents network parameters, $\mathbf{x}=(x,y,z)$ is the 3D coordinates of a sampled point, $\mathbf{d}=(\theta,\phi)$ is the view-in direction, $\mathbf{c}$ is the predicted color, and $\sigma$ is the predicted density. With this scene representation, the rendered color $\mathbf{\hat{C}}(\mathbf{r}(t))$ of the image pixel corresponding to the ray $\mathbf{r}(t)=(\mathbf{o}, \mathbf{d})$ can be evaluated via volume rendering (Eq. \ref{eq:vol_rendering}), where $M$ is the number of samples along $\mathbf{r}(t)$, $\Delta_j$ is the step length of the $j$-th sample, $\tau_j$ can be seen as the probability that the ray can reach the $j$-th sample.
\begin{equation}
    \mathbf{\hat{C}}(\mathbf{r}(t)) = \sum^M_{j=1} \tau_j \mathbf{c}_j,
    \tau_j = \exp\bigg(-\sum_{i=1}^{j-1}\sigma_i \Delta_i \bigg) \bigg(1-\exp\big(-\sigma_j \Delta_j \big)\bigg)
\label{eq:vol_rendering}
\end{equation}
% \begin{equation}
%   \tau_j = exp(-\sum_{t=1}^{j-1}\theta_t \Delta_t)
% \end{equation}
Then, the scene representation can be optimized in a learning fashion via minimizing the photometric loss between the rendered color and the ground-truth color $\mathbf{C}(\mathbf{r}(t))$ in input images (Eq. \ref{eq:nerf_obj_func}). Here $\mathcal{B}$ is a batch of the training rays.
\begin{equation}
  \begin{split}
    \mathcal{L}_{color} &= \sum_{\mathbf{r}(t) \in \mathcal{B}} \left \| \mathbf{\hat{C}}(\mathbf{r}(t)) - \mathbf{C}(\mathbf{r}(t)) \right \|_2^2
  \end{split}
  \label{eq:nerf_obj_func}
\end{equation}
% 5D function is a neural implicit representation (NIR) of any 3D scene.
%
There are also many variants of NeRFs. DirectVoxGO \cite{sun2022direct}, Plenoxels \cite{yu_and_fridovichkeil2021plenoxels}, Instant-NGP \cite{muller2022instant}, TensoRF \cite{chen2022tensorf}, etc. accelerate training of radiance fields by converting the fully implicit scene representations into explicit feature grids with an implicit shading module or spherical harmonics. In particular, TensoRF \cite{chen2022tensorf} factorizes the feature grids into compact low-rank tensor components, achieving considerable rendering quality, shorter training time, and smaller model size. Besides training acceleration, D-NeRF \cite{pumarola2021d} and Nerfies \cite{park2021nerfies} extend NeRF to dynamic scenes via modeling movements as 4D implicit motion fields. Furthermore, HyperNeRF \cite{park2021hypernerf} views each frame of dynamic scenes as a slice in a higher-dimensional latent space and manages to handle topology-varying movements.

% There are also many variants of implicit representation defined for high or lower-dimensional spaces. In higher dimensional spatial-time space, $F(\mathbf{x}, d, t)$ could represent a dynamic scene. While in lower-dimensional 2D space, $F(x,y)$ could be used to compress the image. The implicit representation can be approximated by one or more Multi-Layer Perceptrons (MLPs) sometimes denoted as $F_\theta$. NeRF usually uses volume rendering to produce the image from these colors and densities. The volume rendering is performed by ray marching along the viewing ray. The color and density values are sampled at each step of the ray marching. The color and density values are then combined to produce the final color of the pixel. The volume rendering is differentiable and can be optimized to minimize the photometric loss between the rendered image and the input image.

\subsection{Alpha Compositing}

\begin{figure}[t]
	\centering
	\includegraphics[trim={5 10 5 10},clip,width=\linewidth]{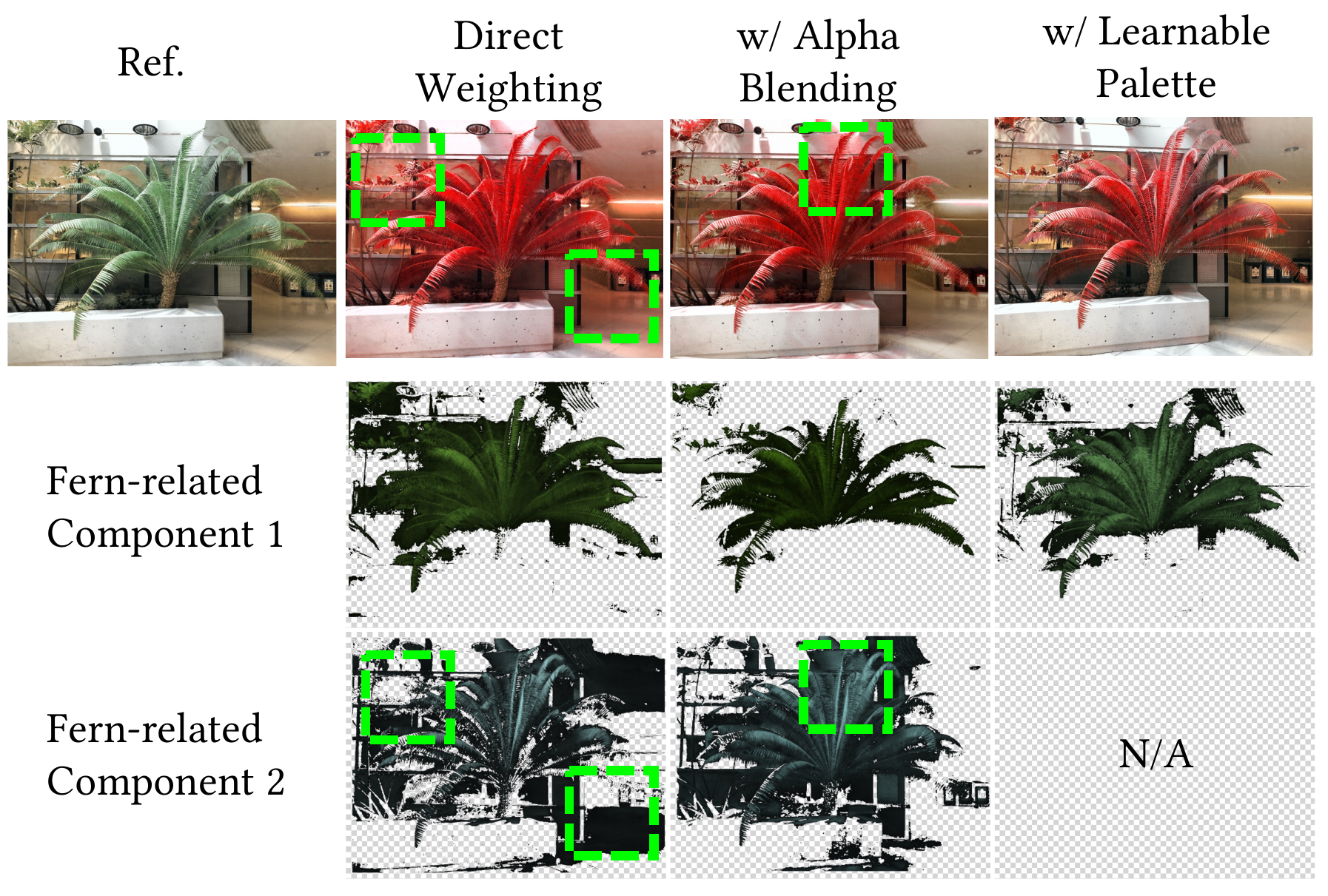}
	%\vspace*{-20px}
	\caption{Ablations on alpha blending and learnable palettes with the goal of recoloring a fern into red. The fern-related layers decomposed via Direct Weight, Alpha Blending and Alpha Blending with Learnable Palette are visualized.}
	\label{fig:ablation_alpha}
	%\vspace*{-10px}
\end{figure}

Alpha compositing, also called alpha blending, is a technique to order-dependently combine multiple image layers into a single image.
% The layers are composited by blending the layers with the alpha channel.
Each image layer is associated with an alpha channel
that is a gray-scale image of values among $[0, 1]$ and specifies the per-pixel opacity of the layer.
% which is usually generated by the user or extracted from the image to blend the layers.
The blending operators are first introduced by \cite{porter1984compositing}.
A commonly-used one is defined as Eq. \ref{eq:over}, where pixel color $C_a$ with opacity $\alpha_a$ is placed over pixel color $C_b$ with opacity $\alpha_b$ to composite new color $C_o$ and opacity $\alpha_o$.
% Among many of the commonly used blending operators, the most basic operation of combining two layers is to put one \textbf{over} the other as described by \cref{eq:over}, which is, in effect, mimics the normal painting operation (e.g. Painter's algorithm):
\begin{equation}
    \alpha_o = \alpha_a + \alpha_b ( 1 - \alpha_a )  ~ , ~
    C_o =  \frac{C_a \alpha_a + C_b \alpha_b ( 1 - \alpha_a )}{\alpha_o}
\label{eq:over}
\end{equation}
%
%
% When alpha compositing is in use, each pixel has an additional opacity recorded in its alpha channel and is expressed as a numeric value between 0 and 1.

To blend multiple image layers, we can recursively apply Eq. \ref{eq:over} to the upper and lower layers. In fact, Eq. \ref{eq:over} depicts a general painting operation. When the opacity of a pixel is 0, there is no overlay on the beneath pixels. While the opacity increases to 1, the pixel becomes opaque and gradually covers the underneath pixels. Our approach extends alpha compositing to 3D scenarios with differentiable properties and view-dependent effects.
% Because the paint compositing operation is a linear blend between two paint colors
% (Eq.\ref{eq:over}), all pixels in the painting lie in the RGB-space convex hull formed by the original paint colors.
% Therefore, any pixel color $p$ can be expressed as the convex combination of the original paint colors $c_i$ as follow:
% \begin{equation}
%   p = \sum w_i c_i, w_i \in [0,1]
% \end{equation}
% where the $w_i$s are generalized barycentric coordinates rather than the opacity values. The generalized barycentric coordinates do not depend on the layer order and $\sum w_i = 1$. They can be converted into layer opacities $\alpha_i$ as follows:
% \begin{equation}
%   \alpha_i = \begin{cases}
%     1 - \frac{\sum_{j=0}^{i-1}w_j}{\sum_{j=0}^{i}w_j} w_i & \text{if } \sum_{j=0}^{i}w_j \neq 0 \\
%     0 & \text{otherwise}
%   \end{cases}
% \end{equation}
% The conversion from generalized barycentric coordinates to opaque layers relies on a predecided order. Due to the division by zero in the second case, the order is undeterminable when a layer is fully opaque(covering everything underneath).
% Due to this ambiguity, we propose to predict layer opacities directly from the implicit representation.

% \begin{figure}[t]
%   \centering
%   \includegraphics[trim={10 0 10 10},clip,width=\linewidth]{fig/23_ARXIV_NETWORK_TENSORF_new_2.pdf}
%   \caption{The architecture of our palette-based layer decomposition radiance fields. We build our underlying radiance field model on the TensoRF \cite{chen2022tensorf} backbone.}
%   \label{fig:network}
%   \vspace*{-10px}
% \end{figure}

\section{RecolorNeRF}

In this section, we describe our proposed method RecolorNeRF, including automatic layer decomposition via alpha blending activation and joint optimization of palettes and scene representations.

\subsection{Palette-based Layer Decomposition via Alpha Blending Activation}

Suppose we have an ordered palette $P=\big\{p_i \in [0, 1]^3\big\}_{i=1}^{K}$, we aim to decompose a radiance field into $K$ pure colored radiance layers, where the ith layer corresponds to the ith item of the palette. We define $p_1$ as the color of the background layer and $p_K$ as the color of the topmost layer. Our scene representation is modeled as Eq. \ref{eq:recolornerf_func}.
\begin{equation}
\label{eq:recolornerf_func}
F_{\Theta}(\mathbf{x}, \mathbf{d}) \rightarrow \left (\alpha_1, \alpha_2, \dots, \alpha_K, \sigma \right)
\end{equation}
The model computes view-dependent opacity values $\alpha_i(\mathbf{x},\mathbf{d})$ for each layer. After retrieving opacity values from the model, we composite the radiance layers via alpha blending to yield the radiance field. By recursively applying Eq. \ref{eq:over} to the opacity fields $\alpha_i(\mathbf{x},\mathbf{d})$ and the palette $P$, we can obtain $\mathbf{c}(\mathbf{x},\mathbf{d})$ as Eq. \ref{eq:alpha}:
\begin{equation}
  \mathbf{c}(\mathbf{x},\mathbf{d}) = p_K + \sum_{i=1}^{K-1} \left [ (p_{i} - p_{i+1}) \prod_{j={i+1}}^{K-1} ( 1 - \alpha_j(\mathbf{x},\mathbf{d}) ) \right ]
  \label{eq:alpha}
\end{equation}
The cumulative multiplication in Eq.\ref{eq:alpha} will cause optimization numerically unstable. To better integrate alpha compositing into deep models as a stable differentiable function, we sort out coefficients of each $p_i$ and transform them into logarithmic space:
\begin{equation}
  \mathcal{T}(\alpha_i(\mathbf{x},\mathbf{d})) = \begin{cases} \sum_{j=i+1}^K \log (1-\alpha_j(\mathbf{x},\mathbf{d})) & i=1 \\
    \log \alpha_i(\mathbf{x},\mathbf{d}) + \sum_{j=i+1}^K \log (1-\alpha_j(\mathbf{x},\mathbf{d}))  & 1<i<K \\
    \log (1-\alpha_i(\mathbf{x},\mathbf{d})) & i=K \end{cases}
  \label{eq:barycentric}
\end{equation}
% \begin{equation}
%   \mathcal{T}(\alpha_i(\mathbf{x},\mathbf{d})) = \begin{cases}
%       \log (\alpha_i(\mathbf{x},\mathbf{d})) & i=1 \\
%         \log \alpha_i(\mathbf{x},\mathbf{d}) + \sum_{j=i+1}^K \log (1-\alpha_j(\mathbf{x},\mathbf{d}))  & 1<i<K \\
%         \log (\alpha_i(\mathbf{x},\mathbf{d})) & i=K
%     \end{cases}
%   \label{eq:barycentric}
% \end{equation}
%
% Generalized barycentric coordinates express any point $p$ inside a polyhedron as a weighted average of the polyhedron's vertices $c_i$.
We can show that $\sum_{i=1}^K \exp
\big[ \mathcal{T}(\alpha_i(\mathbf{x},\mathbf{d}))\big] = 1$. Thereby we define an alpha blending activation $\zeta(\cdot)$ as Eq. \ref{eq:abact}, which maps opacity to generalized barycentric coordinates. This alpha blending activation is differentiable and numerically stable, compared with the cumulative multiplication in Eq. \ref{eq:alpha}. Consequently, the radiance field can be seen as a convex combination of the palette $p_i \in P$:
\begin{equation}
    \mathbf{c}(\mathbf{x},\mathbf{d}) = \sum_{i=1}^K \zeta(\alpha_{i}(\mathbf{x},\mathbf{d})) p_i
    , ~ \text{where} ~
    \zeta(\alpha) = \exp
    \big[ \mathcal{T}(\alpha)\big]
    \label{eq:abact}
\end{equation}
% This nice convex property of the blending weights cancels the need for softmax-like group activation.

Since the geometry of the scene is independent of the opacity values, we separate the parameters of the geometry and appearance models. To achieve this, we adopt TensoRF \cite{chen2022tensorf} as our backbone, which simultaneously enables fast model training. The overview pipeline of our method is present in Fig. \ref{fig:network}.

\subsection{Palette Learning}

\begin{figure}[t!]
	\centering
	\includegraphics[trim={0 10 0 10},clip,width=\linewidth]{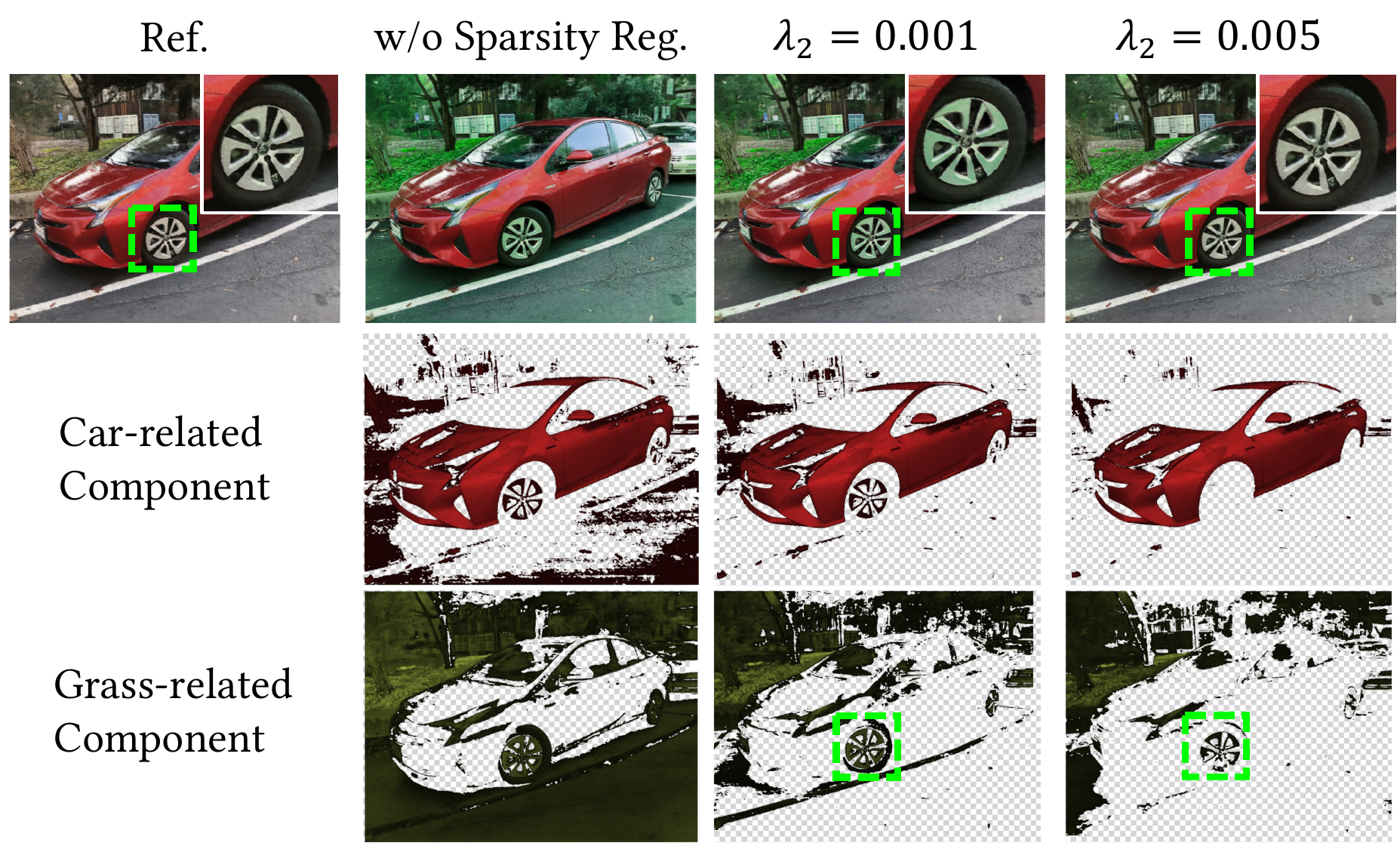}
	\vspace*{-16px}
	\caption{Ablations of sparsity regularization with the goal of recoloring the background grass into greener. The component of foreground car and background grass are visualized.}
	\label{fig:ablation_sparsity}
	\vspace*{-8px}
\end{figure}

% Automatic layer decomposition relies on the color palette of the image.
The key to successful layer decomposition is an optimal palette $P$ that can express the entire color space of the 3D scene. For 2D images, \citet{tan2016decomposing} propose to model a palette as convex hull vertices of the pixel colors. In this way, any pixel color can be uniquely expressed as a convex combination of the palette colors.
Inspired by this, we choose to formulate $P$ as a set of convex hull vertices, encompassing the entire color space of the scene.
Note that our alpha blending activation can  coherently support this formulation since it naturally yields convex combinations of the palette.
However, directly using the convex hull vertices as palettes is not feasible for user editing due to the convex hull will have thousands of vertices. \citet{tan2018efficient} design an iterative heuristic convex hull simplification algorithm for reducing the palette to a user-desired size.
Our empirical study finds that palettes pre-computed from this heuristic approach lead to many failure cases of layer decomposition in 3D scenes since only the 2D image-level color distribution is exploited. In this regard, we incorporate learnable palettes into the optimization of 3D scene representation and layer decomposition, as a group of training parameters. In order to preserve the geometric properties of palettes, we propose novel \textit{convex hull bounding } and \textit{projection-based convex} regularizations, which jointly guide the optimization trajectory of the palette.
% Specifically, we model palette learning as a vertex search problem on a larger prior convex hull $U$.

\paragraph{Convex Hull Bounding Regularization}
%
% Tan et al. \cite{tan2016decomposing} proposed to compute vertices on the convex hull of the pixel colors based on the observation that the color distributions from paintings and natural images take on a convex shape in RGB space.
%
% Since the convex hull tightly wraps the observed colors, its vertex colors can be blended with a convex combination to reproduce any color in the image and be used as the palette.
%
% They also propose an iterative simplification algorithm to reduce the palette to a user-desired size. After simplification, the remaining colors in the palette are the representative colors in the image suitable to decompose the image into layers.
%
% Inspired by the above observation, we use the convex hull to form a feasible region to constrain the choice of colors in the palette of the RecolorNeRF. The convex hull $U$ is then given by the expression:
%
The learnable palette $P$ should not be unbounded, since $P$ is a subset of $[0,1]^3$ at most. To further shrink the bound of palette colors, we build a bounding convex hull from all input images (Eq. \ref{eq:pltbd}).
\begin{equation}
U = \bigg\{ \sum_{c_j\in\mathcal{P}} \lambda_j c_j ~|~ \lambda_j \geq 0 ~\text{for all}~ j ~\text{and}~ \sum_{j} \lambda_j = 1 \bigg\}
\label{eq:pltbd}
\end{equation}
where $\mathcal{P}$ is the universal set of all pixel colors in all multi-view images.
We employ this bounding convex hull to form a feasible region by penalizing out-of-bound palette colors. To achieve this, we first find a set $S(U)$ of all simplices of the bounding convex hull $U$. Then, we regularize the Euclidean distance from palette color $p_i$ to its closest point on the simplex facets $s \in S(U)$, if $p_i \notin U$. The described regularization term is given in Eq. \ref{eq:bdloss}.
% and $Nearest(c|s)$ is the closest point on the simplex facet $s$ to the color $c$.
\begin{equation}
  \mathcal{L}_{bd} = \sum_{i=1}^K \omega_{out}~ \vmathbb{1}{\{p_i\notin U\}} \min_{ s \in S(U)} \big\|p_i - \text{Nearest}(p_i|s) \big\|_2
  \label{eq:bdloss}
\end{equation}
where $\omega_{out}$ is the penalty strength, $\text{Nearest}(p_i|s)$ finds closest point on the simplex facet $s$ to the color $p_i$, and $\vmathbb{1}{\{p_i\notin U\}}$ denotes the indicator of $p_i\notin U$.

\paragraph{Projection-based Convex Regularization}

In addition to bounding the range of palette colors, the palette is expected to maintain its convexity. Our solution is to project the palette colors to the vertices of the bounding convex hull. In essence, this projection-based optimization strategy imposes a constraint $P \subset \text{HullVertices}(U)$. Specifically, the regularizer in Eq. \ref{eq:projloss} is proposed to turn palette gradients towards the vertices of $U$.
\begin{equation}
  \mathcal{L}_{proj} = \sum_{i=1}^K \omega_{in}~ \vmathbb{1}{\{p_i\in U\}} \min_{ v \in \text{HullVertices}(U)} \big\|p_i - v \big\|_2
  \label{eq:projloss}
\end{equation}

Finally, we combine the two regularizations to $\mathcal{L}_{hull} = \mathcal{L}_{bd} + \mathcal{L}_{proj}$. In our implementation, we generally set $\omega_{out}= 1$ and $\omega_{in}= 0.001$ to enforce a strong out-of-bound penalty and slow down the projection to the bounding convex hull, in order to prevent palette colors from fast convergence to nearby sub-optimal vertices.

% \subsection{Alpha Blending Activation}
% We composite the layers fields to produce the radiance field and compute the final color $C(r)$. This stage takes the opacity fields $\alpha_i(x,d)$ and the palette $P=\{c_0, c_1, \cdots, c_K\}$ as input and outputs the color field $r(x,d)$:
% %
% \begin{equation}
%   r_j = c_K + \sum_{i=1}^K \left [ (c_{i-1} - c_i) \prod_{j=i}^K ( 1 - \alpha_j  ) \right ]
%   \label{eq:alpha}
% \end{equation}
% %
% where the background color $c_0$ is opaque therefore $\alpha_0$ is set to constant value 1. The cumulative multiplication in Eq.\ref{eq:alpha} makes the optimization of the opacity fields not numerically stable. To avoid this problem, we reformulate the alpha composition in logarithmic space by converting the opacity to generalized barycentric coordinates:
% %
% \begin{equation}
%   w_i = \begin{cases} \sum_{j=i+1}^K \log (1-\alpha_j) & i=0 \\
%     \log \alpha_i + \sum_{j=i+1}^K \log (1-\alpha_j)  & 0<i<K \\
%     \log (1-\alpha_i) & i=K \end{cases}
%   \label{eq:barycentric}
% \end{equation}
% %
% Generalized barycentric coordinates express any point $p$ inside a polyhedron as a weighted average of the polyhedron's vertices $c_i$. We can show that $\sum_{i=1}^K exp(w_i) = 1$ and $r_j = \sum_{i=1}^K exp(w_{ij}) c_i$. This nice convex property eliminates the need for Softmax-like group activation. We use Sigmoid followed by Eq.\ref{eq:barycentric} as the output activation of opaque and name it \textit{alpha blending activation}. The alpha blending activation is differentiable and numerically stable compared to cumulative multiplication.

\paragraph{Palette Initialization}

As learnable parameters, random initialization of the palette could work for most cases. However, a user-designed prior palette can further promote the editability of the scene. For instance, one may move his/her target edit layer to the top layer and drop undesired layers to the background, in accordance with the alpha blending property that the color of the upper layer relates to fewer areas/objects. To this end, we support a human-involved palette initialization strategy, where the user first inputs a prior palette with expected ordering, then our method jointly optimizes the input palette to attain better layer decomposition. We find prior palettes can facilitate our method to extract representative layers in more complex scenes.

% initialization counts for the
% The ``over'' operator we employ to combine the layers is order-dependent, i.e., not commutative. Therefore, the order of the layer arrangement matters. For $K$ layers, there are $K!$ different orderings. It is inefficient to try each and re-train the model with all possible orderings. Hence, we summarize 3 empirical solutions to   We assign pixels in multi-view images to their most contributed layer by computing the L2 distance with the corresponding palette color. Then we sort layers by their pixel counts and use the layer with the smallest pixel count as the topmost layer as we find this arrangement usually decomposes sparser opacity. Other criteria such as the total opacity, gradient of opacity, or Laplacian of opacity could also be used based on the specialty of the scene. The user can also directly assign the order according to human preference. Note that the last layer is always a non-transparent layer corresponding to the background color.

\subsection{Rendering and Optimization}

\begin{figure}[t]
	\centering
	\includegraphics[trim={0 10 0 20},clip,width=\linewidth]{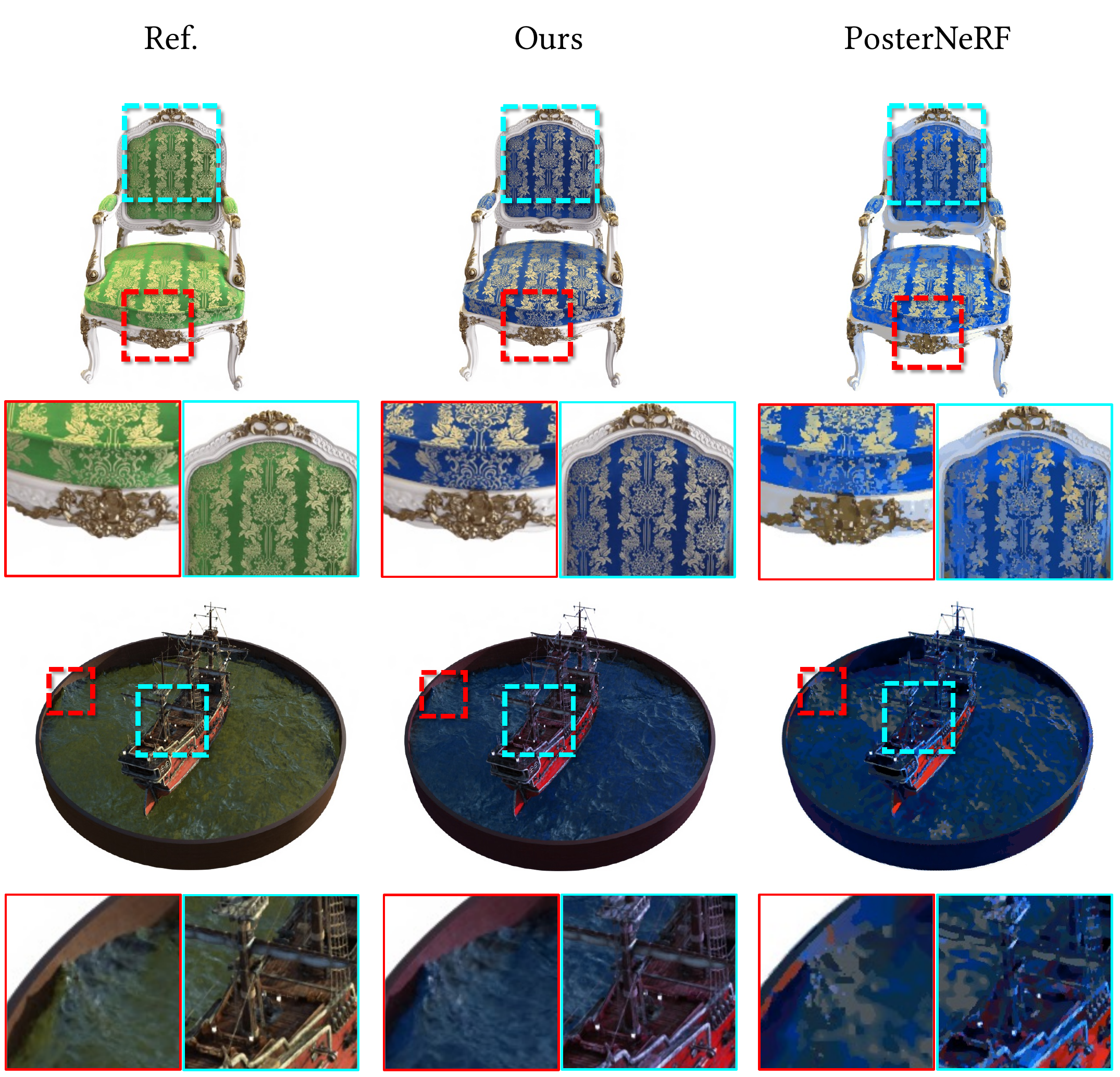}
	% \includegraphics[width=\linewidth]{fig/res_vis_cmp2_2.pdf}
	%\vspace*{-18px}
	\caption{Comparisons of our method and PosterNeRF on Synthetic NeRF dataset. In the first case, we aim to recolor the green chair into the blue without change of the yellow patterns. In the second case, we aim to recolor the orange ship into red  and tune the water surface into blue.}
	\label{fig:cmp-our-posternerf}
	%\vspace*{-10px}
\end{figure}

% We render the composited layers for every batch of training rays and optimize rendering loss to reconstruct the scene representation.
The volumetric rendering with the alpha compositing for a training ray $\mathbf{r}(t)=(\mathbf{o}, \mathbf{d})$ is given in Eq. \ref{eq:vol_rendering_alpha}.
\begin{equation}
    \mathbf{\hat{C}}(\mathbf{r}(t)) = \sum^M_{j=1} \tau_j \sum_{i=1}^K \zeta(\alpha_{i}(\mathbf{x}_j,\mathbf{d})) p_i
  \label{eq:vol_rendering_alpha}
\end{equation}
Recall that $M$ is the number of samples along the ray and $\tau_j$ is the accumulated transmittance, as described in Eq. \ref{eq:vol_rendering}. The scene representation can be optimized by minimizing the squared errors between the rendered color and the ground truth color, as in Eq. \ref{eq:nerf_obj_func}. To encourage sparser layer decomposition, we add a \textit{layer sparsity} regularization to the scene optimization.

% We render the composited layers and optimize the NeRF model with color loss. Once the NeRF model is optimized, We can render the recolored NeRF model to generate photo-realistic images. The rendering is differentiable with the following equation:
%
% \begin{equation}
% 	\mathbf{C(r)} = \sum_M \tau_j(1-exp(-\theta_j \Delta_j)) r_j
%   \label{eq:rendering}
% \end{equation}
%
% and  $\tau_j$  is the accumulated transmittance, representing the probability that the ray travels from $t_1$ to $t_j$ without being intercepted, given by:
%
% \begin{equation}
%   \tau_j = exp(-\sum_{t=1}^{j-1}\theta_t \Delta_t)
% \end{equation}
% where $\Delta_t = t_{i+1} - t_i$ is the distance between adjacent samples. This function
% for calculating $\mathbf{C(r)}$ from the set of $(w_i, \theta_i )$ values is trivially differentiable for both palette and opacity. The optimization is supervised by the total squared error between the rendered and true pixel colors $\mathbf{C(r)} $ and $\mathbf{C}_{gt}$:
%
% \begin{equation}
%   \begin{split}
%     \mathcal{L}_{color} &= \sum_{x,d} \left \| \mathbf{C(r)}  - \mathbf{C}_{gt} \right \|_2^2
%   \end{split}
% \end{equation}
% %
% Note that $r$ is the ray shot from the camera. The optimization is performed by minimizing the color loss $\mathcal{L}_{color}$ with respect to the NeRF model parameters and the palette $P$.

\paragraph{Layer Sparsity Regularization}

The sparsity of opacity fields is preferred during layer decomposition. Sparser weights of the palette indicate the palette is more representative and the decomposition is more complete. We first render a weighting map $W_i(\mathbf{r}(t))$ for each layer to estimate palette components around occupied space:
\begin{equation}
	W_i(\mathbf{r}(t)) = \sum^M_{j=1} \tau_j \zeta(\alpha_{i}(\mathbf{x}_j,\mathbf{d}))
\end{equation}
Next, we aim to penalize the L0 norm of these components, i.e., $\lVert [W_1, \dots, W_K] \rVert_0$. Since the L0 norm is not differentiable, we design a specialized soft counting norm (Eq. \ref{eq:softl0}) based on the Sigmoid function to conduct the sparsity penalty.
\begin{equation}
  \begin{split}
    h_{0}(\mathbf{W}) &= \sum_{l=1}^K \exp{\bigg(\frac{l}{K} - 1\bigg)} \bigg[1 + \exp{(-\eta W_l + \beta)}\bigg]^{-1}
  \end{split}
  \label{eq:softl0}
\end{equation}
The first exponential term in $h_0$ order-dependently re-scales the sparsity of different layers such that the upper layers gain more sparsity than the lower layers. The reciprocal term in $h_0$ approximates the Heaviside function on $[0, 1]$, where $\eta$ and $\beta$ are sparsity strength hyper-parameters, controlling the step edge sharpness and counting threshold. In our implementation, $\beta$ is constantly set to 6 and $\eta$ is set to 24 or 48, depending on the scene's complexity. In the end, the batch-wise sparsity regularization is written as Eq. \ref{eq:spsloss}.
\begin{equation}
  \begin{split}
    \mathcal{L}_{sparsity} &= \sum_{\mathbf{r}(t) \in \mathcal{B}} h_{0}\big([W_1(\mathbf{r}(t)), \dots, W_K(\mathbf{r}(t))]\big)
  \end{split}
  \label{eq:spsloss}
\end{equation}

%
% \begin{equation}
%   \begin{split}
%     \mathcal{L}_{sparsity} &= \sum_{x,d} - \left \| 1 - \mathbf{E}_{opqaue}(\mathbf{r}) \right \|_2
%   \end{split}
% \end{equation}
% \begin{equation}
%   \begin{split}
%     \mathcal{L}_{sparsity} &= \sum_{x,d} \left \| \frac{1}{1 + \exp{(-\eta\mathbf{E}_{opqaue}(\mathbf{r}) + C_0)}} \right \|_1
%   \end{split}
% \end{equation}
%
% where $\eta$ is a hyper-parameter standing for the sparsity strength. Since L0 sparsity norm is not differentiable, we alternatively adopt a scaled and shifted (by $C_0$) sigmoid function to mimic the stepping shape of the L0 norm among $[0, 1]$.
%
% Intuitively, negatively squaring $1-\alpha_i$ produces an
% the objective function that would ``prefer'' to increase $\alpha_i$ away from 1, where the farthest value from 1 is 0, resulting in a sparse solution.
% This unusual formulation is possible because the $\alpha_i$ s are bounded in the interval $[0, 1]$ by Sigmoid.
% Our first regularization term penalizes opacity; absent additional information, a completely occluded layer should be transparent.
% Note that the transmittance weighting $\tau_j$ in $\mathbf{E}_{opaque}$ is vitally important to the sparsity of layer decomposition, and the reason is to be investigated. This is left as future work and it remains as a magic of volumetric rendering at least for now.

\paragraph{Overall Loss}

The overall objective function of our joint optimization is the weighted sum of photometric reconstruction loss, palette regularization $\mathcal{L}_{hull}$ and sparsity penalty $\mathcal{L}_{sparsity}$, defined as:
\begin{equation}
    \mathcal{L} = \mathcal{L}_{color} + \lambda_{1} \mathcal{L}_{hull} + \lambda_{2} \mathcal{L}_{sparsity}
\end{equation}
where $\lambda_{1} = 1$ and $\lambda_{2}=0.001$ or $0.005$ in our experiments.

\paragraph{Recoloring}

%Upon successful decomposition of the scene into various layers with the learned palette, RecolorNeRF allows user manipulation of layer color by adjusting palette components.
% While RecolorNeRF represents colors in the RGB space, users have the flexibility to employ any color space for modification, including but not limited to CIELAB, HSV, and CMYK.
% The recoloring process is then done by re-rendering the scene with the adjusted palette. With typically less than 10 palettes per scene, users can efficiently edit diverse color schemes.

Upon decomposing a scene into layers with a learned palette, RecolorNeRF enables user color manipulation via palette adjustment, allowing efficient editing of diverse color schemes with typically less than 10 palettes per scene through re-rendering.

\begin{figure*}[t]
	\centering
	\includegraphics[trim={0 10 0 5},clip,width=\textwidth]{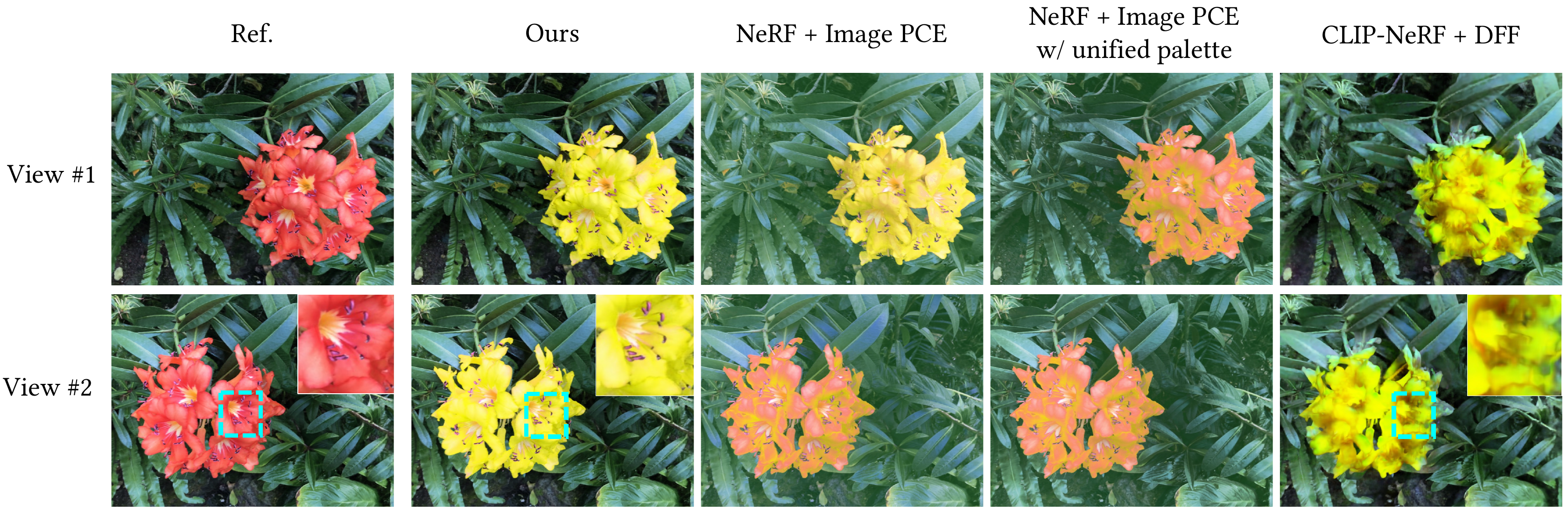}
	\vspace*{-18px}
	\caption{Comparisons of our method, NeRF + Image PCE, NeRF + Image PCE w/ unified palette and CLIP-NeRF + DFF. We aim to recolor the red flower into yellow. For our method and PCE baselines, we replace red components with yellow to perform recoloring. For CLIP-NeRF + DFF, we fine-tune an optimized DFF model with text prompt `yellow flower' and background filtered out.}
	\label{fig:cmp-llff}
	\vspace*{-12px}
\end{figure*}

\section{Experimental Results}

% In this section, we present our experimental results.

% We implemented our full model based on TensoRF \cite{chen2022tensorf}, considering its fast training speed and superior rendering quality.
% We first compare the recoloring results of our method with a baseline method PosterNeRF \cite{tojo2022recolorable}. Then we will present an ablation study to validate the effectiveness of our proposed modeling and regularization. We also showcase multiple recoloring results of our method both on complex scenes and objects.

% We first introduce the dataset and evaluation metrics. Then, we present the quantitative and qualitative results of the proposed method.

\subsection{Evaluation}

\paragraph{Datasets} We evaluate our method on 360$^\circ$ synthetic objects (Synthetic NeRF \cite{mildenhall2021nerf} and Synthetic NSVF \cite{liu2020neural} datasets), forward-facing real-world scenes (LLFF \cite{mildenhall2019local}), and 360$^\circ$ real-world objects (Tank and Temples\cite{liu2020neural,knapitsch2017tanks}).

\vspace{-0.1in}
\paragraph{Baselines} We compare our method with 4 recent state-of-the-art image-level or NeRF-level color editing baselines. For image-level PCE, we apply Tan et al.'s approach \cite{tan2018efficient} to NeRF rendering results. We consider 2 schemes for building palettes for novel views. 1) Each novel view admits an independent palette. We dub this scheme \textbf{NeRF + Image PCE}. 2) Novel views share a unified palette built from all views' pixels. We dub this one \textbf{NeRF + Image PCE w/ unified palette}. In addition to image-level editing, \textbf{PosterNeRF} \cite{tojo2022recolorable} is included as a NeRF-level PCE baseline on the Synthetic NeRF datasets. We also compare with the recent text-driven scene editing method CLIP-NeRF \cite{wang2022clip}. To be precise, we use its follow-up implementation with distilled feature fields (DFF) \cite{kobayashi2022distilledfeaturefields}, dubbed \textbf{CLIP-NeRF + DFF}.
% Hyper-parameters of these baselines are in accordance with their official implementations.

\begin{figure}[t]
  \centering
  \includegraphics[trim={0 10 0 0},clip,width=\linewidth]{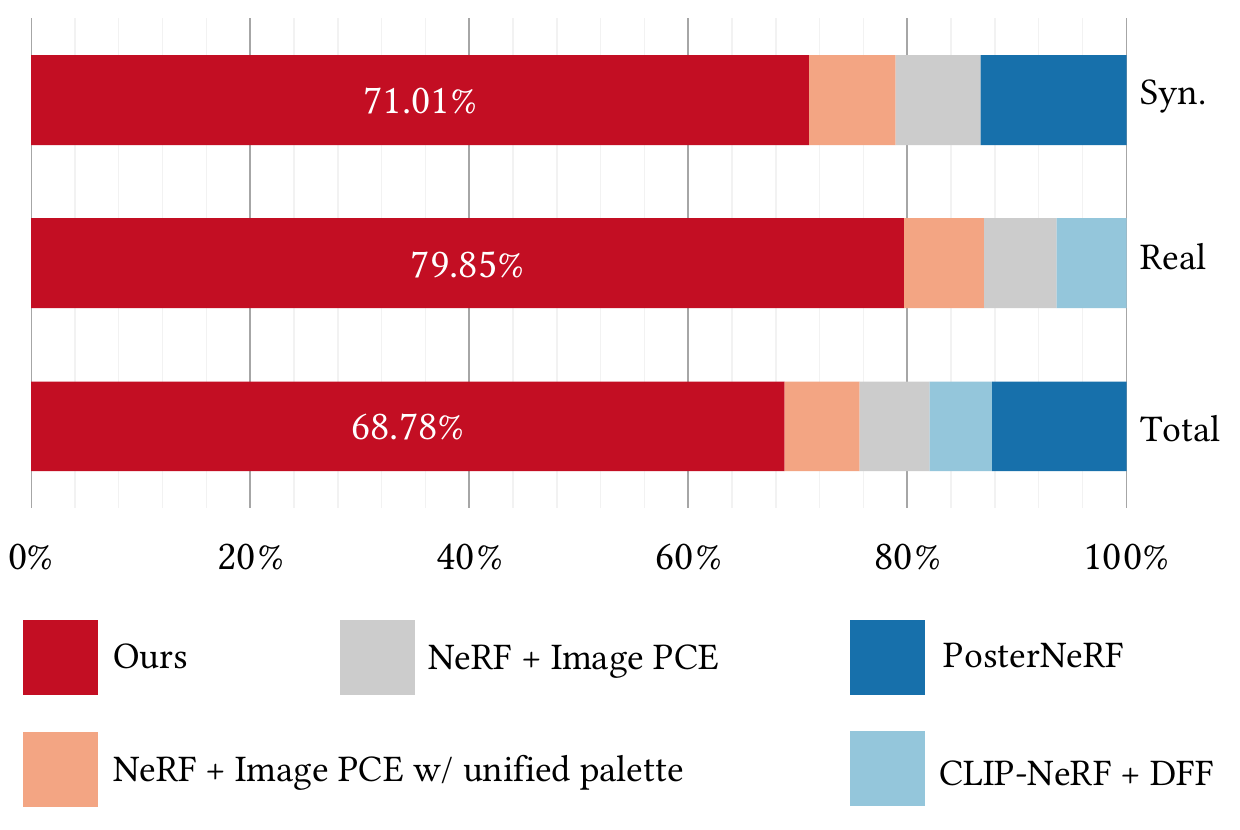}
  \vspace*{-16px}
  \caption{User study statistics. Each color bar of a corresponding method indicates the user selection proportion among all questions that compare the results of the method.
  In the synthetic scenes, the proportion is 71.01\%/7.91\%/7.76\%/13.33\% from left to right.
  In the real-world scenes, the proportion from left to right is 79.85\%/7.32\%/6.66\%/6.37\%.
  In total, the proportion is
  68.78\%/6.86\%/6.41\%/6.58\%/5.70\%.
  % The first row shows the percentage of users' selection of different methods on the synthetic scenes. The second row shows the user selection on real-world scenes. The third row shows the total percentage of both synthetic and real-world scenes.
  }
  \label{fig:userstudy}
  \vspace*{-8px}
\end{figure}

% \vspace{-0.1in}
\paragraph{User Study}
% Given that color correctness and consistency is very objective to artists,
To quantitatively evaluate recoloring performance from users' aesthetic perspective, we conduct a user study to validate the advantage of our method. We first prepare 25 editing targets for various scenes (14 synthetic objects and 11 real-world scenes). Then, we accordingly generate recoloring results using our method and other baselines (only if the baseline supports the case). For each case, we ask respondents to select the most satisfied recoloring result that matches the editing target best. Our user study questionnaires are distributed online with randomized orders of questions and choices. Respondents are encouraged to spend at least 20 minutes completing the questionnaire. We collect 5081 valid ratings from 68 independent respondents.
A visualization of the user study results is shown in Fig. \ref{fig:userstudy}.
%
%
% We analyze the results by
% The 25 cases are divided into two categories: synthetic and real-world. 14 of them are editing real-world scenes and the rest are editing synthetic objects.
We can see that our method outperforms the other baselines both in real-world
% (82.44\%)
and synthetic
% (75.44\%)
scenes by a large margin. Especially for real-world scenes with complex color distributions, our method is much preferred.
% Our alpha blending with learnable palettes shows more advantages on the complex scenes.
The overall results reflect our method is able to generate more aesthetic editing on various scenarios and possesses better recoloring capacity.

% \vspace{-0.1in}
\paragraph{Qualitative Comparison}
In qualitative evaluation, we visualize results from different methods and compare their rendering quality and recoloring capability. Fig. \ref{fig:cmp-our-posternerf} exhibits the comparison results of our method versus PosterNeRF on the Synthetic NeRF datasets. From the close-ups, we can observe our method outperforms PosterNeRF both in rendering quality and recoloring. Our method can precisely extract editing patterns and recolor them without polluting other areas. In contrast, PosterNeRF blends the colors of editing areas and undesired areas. . Fig. \ref{fig:cmp-llff} compares our method with other 3 baselines in a forward-facing ``flower'' scene. We can observe that view consistency is not guaranteed by NeRF + Image PCE. Even if using a view-shared palette, NeRF + Image PCE w/ unified palette fails to recolor the red flower but mistakenly brightens the background leaves. CLIP-NeRF + DFF manages to change the color of the flower by text prompt. However, its rendering quality is compromised. Also, CLIP-NeRF + DFF is limited by the expressions of text prompts when it is difficult to describe editing targets in text.
% In the first case, our method can extract the editing components, while PosterNeRF fails to bypass those delicate patterns. In the second edit, the water surface is recolored by our method without polluting the ship's color. In contrast, PosterNeRF blends the colors of the ship and the water surface.

\begin{figure*}[t]
	\centering
	\includegraphics[trim={0 12 0 20},clip,width=\textwidth]{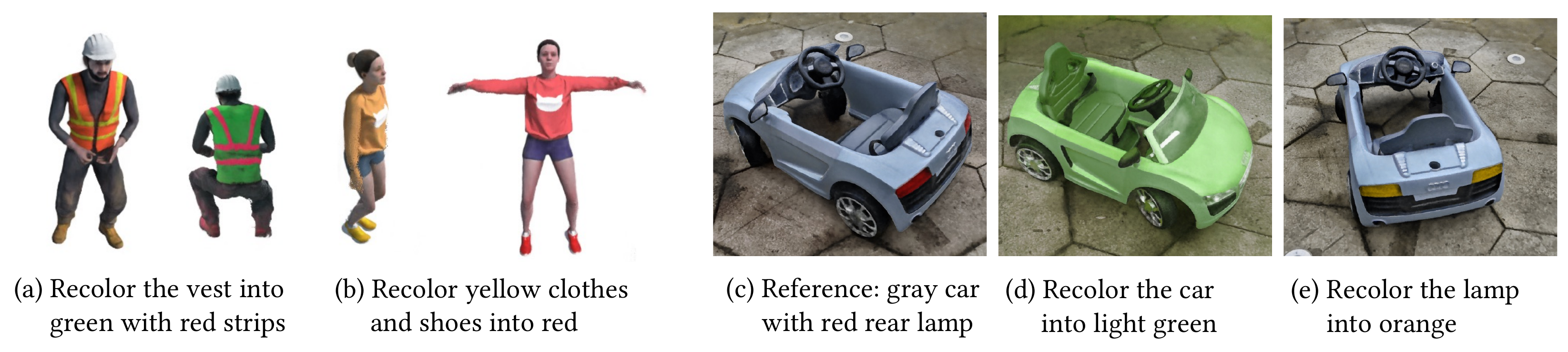}
	%\vspace*{-18px}
	\caption{Recoloring results of our method using D-NeRF \cite{pumarola2021d} and Ref-NeRF \cite{verbin2022ref} as backbones. (a) and (b) demonstrates recoloring of synthetic dynamic objects. (c) (d) and (e) showcase recoloring of real-world 360$^{\circ}$ scenes with metallic and translucent materials.}
	\label{fig:dnerf-refnerf-backbones}
	%\vspace*{-15px}
\end{figure*}

\subsection{Ablation Study}

% We evaluate the layer decomposition with and without sparsity regulation, alpha blending, and learnable palette. The evaluation is based on the TensoRF backbones. We use the same setting in comparison experiments. The effect of sparsity regulation is shown in Fig.\ref{fig:ablation_sparsity} and the effects of alpha blending and learnable palette are shown in Fig.\ref{fig:ablation_alpha}.

\paragraph{Sparsity Regularization} We do an ablation study to certify the effect of sparsity regularization. Fig. \ref{fig:ablation_sparsity} visualizes layers of car and grass, decomposed under no sparsity regularization, weak sparsity regularization ($\lambda_2 = 0.001$) and strong sparsity regularization ($\lambda_2 = 0.001$), respectively. Without any regularization on sparsity, the grass component is dispersed in the entire background, leading to obvious color pollution when recolored into greener. By adding weak regularization on layer sparsity, the car layer and grass layer are separated out. Looking into the recoloring result, we can find slight color pollution occurs on the tire of the car. By imposing stronger sparsity regularization, the decomposition becomes more complete and further mitigates the color pollution.

% we compare the layer decomposition results with and without sparsity regularization. We also show the results under different penalty weights $\lambda_{2}$. Without sparsity regularization ($\lambda_{2}=0$), the decomposed layer are overlapped and the same pixels are associated with multiple layers. With the stronger sparsity regularization ($0.001 \le \omega_{L0} \le 0.01$), the decomposed layers are more separated and sparser. The sparsity regularization can also help RecolorNeRF learn a more representative palette by dropping redundant layers and adjusting colors in the palette. If the sparsity regularization is too strong ($\omega_{L0} \ge 0.1$), the sparsity penalty $\mathcal{L}_{sparsity}$ will dominate the color loss $\mathcal{L}_{color}$ and most of the decomposed layers will be dropped. Without enough colors, the reconstruction will fail.

\paragraph{Alpha Blending with Learnable Palette} To evaluate the effect of alpha blending, we replace the alpha blending with the direct weighting of the palette (named Direct Weighting). We also disable the learnable palettes to evaluate the effect of our palette learning. As shown in Fig. \ref{fig:ablation_alpha}, we can see  Direct Weighting scheme fails to completely decompose the foreground and the background even with our sparsity regularization. With this deficiency, recoloring the foreground will pollute the tones of the background. Alpha blending facilitates layer decomposition and alleviates the color pollution problem during recoloring. We owe this to the order-dependent formulation in the alpha blending. Furthermore, the learnable palette can further promote the separation of the background and foreground layers. In the shown case, the fern-related components are optimized to be merged. Thus, the green component becomes more representative, leading to better recoloring results.
% We can see that decomposition by our RecolorNeRF (last row) is the most disjoint and representative, which enables color editing more precise and effective.

% \begin{figure*}[t]
%   \centering
%   \includegraphics[width=\textwidth]{fig/res_vis1_1.pdf}
%   \caption{Gallery of our color editing results. The first image of each row is the reference before editing along with the beneath optimized palette. The other 3 images showcase 3 examples of color editing with the corresponding edited palettes.}
%   \label{fig:res_vis1}
% \end{figure*}

\paragraph{Synthesis Quality with Palette}
To assess the quality of our radiance field rendering using alpha blending and palettes, we compare photometric errors of our method and the original backbone TensoRF. The palettes in our model are initialized with 2 schemes: 1) user input and 2) automatic palette extraction via RGB-space hull simplification \cite{tan2016decomposing}. Table \ref{tab:psnr} reports the PSNR, SSIM, and LPIPS metrics. The user-customized palette scheme offers more controllable color editing, albeit with slightly lower performance. Our model initialized with automatically extracted palettes yields even better rendering results, further validating the effectiveness of alpha blending with palettes in view synthesis.
% We evaluated the quality of our radiance field rendering, which utilizes alpha blending and palettes, by comparing the photometric errors of our method and the original backbone TensoRF. We initialize the palettes in our model using two schemes: 1) user input and 2) automatic palette extraction via RGB-space hull simplification \cite{tan2016decomposing}. Table \ref{tab:psnr} presents the PSNR, SSIM, and LPIPS metrics. The user-customized palette scheme offers more controllable color editing, albeit with slightly lower performance. Our model initialized with automatically extracted palettes yields even better rendering results, further validating the effectiveness of alpha blending with palettes in view synthesis.

% To evaluate the effect of using additional palette, we compare the quality of novel view synthesis result with original backbone. The RecolorNeRF are optimized with both alpha blending and learnable palette. In order to compare with the original backbone, the learned palette are kept unchanged to generate the exactly same image. The ablative metrics are reported below. As found, using palettes does not cause obvious quality reduction.

\begin{table}[]
    % \footnotesize
    \caption{Comparisons of novel view synthesis results between the original backbone and RecolorNeRF initialized with user-input palettes (user init.) and automatic palettes via hull simplification (auto init.), respectively.}
    \begin{tabular}{l|l|l|l}
        \hline
        Models & PSNR $\uparrow$   & SSIM $\uparrow$  & LPIPS $\downarrow$ \\ \hline
        Original TensoRF       & $28.32 \pm 5.21$ & $0.852 \pm 0.119$ & $0.186 \pm 0.129$ \\ \hline\hline
        Ours (user init.)      & $28.03 \pm  4.87$ & $0.843 \pm 0.125$ & $0.199 \pm 0.138$ \\ \hline
        Ours (auto init.) & $28.76 \pm  4.86$ & $0.869 \pm 0.111$ & $0.169 \pm 0.128$ \\ \hline
    \end{tabular}
    \label{tab:psnr}
    % \vspace{-0.2in}
\end{table}

\subsection{More Visual Results}

We showcase the results of our RecolorNeRF in Fig. \ref{fig:res_vis1} and more in supplementary materials.
Moreover, our approach can universally support other backbones. Fig. \ref{fig:dnerf-refnerf-backbones} shows the results of our approach with D-NeRF \cite{pumarola2021d} and Ref-NeRF \cite{verbin2022ref} backbones. This also illustrates the prospective applications of our method, e.g., virtual car repainting and digital clothes recoloring in online shopping.

% Our RecolorNeRF can generate photo-realistic free-view images in challenging scenarios. In the first ``Ficus'' scene, our method can recolor the small twigs to ``rosewood'' (the last column). Moreover, the specular lighting components are editable by our method, as shown in the last column of the ``Drums'' scene. The ``Truck'' scene validates the recoloring performance of our method on a real 360$^{\circ}$ scene. In addition to recoloring foreground objects, the last column of the ``Fern'' and the third column of the ``Orchids'' scenes illustrate the effects of background editing by our method. We include more visualization results in the supplementary material.

\section{Conclusion}

RecolorNeRF is an effective method to generate recolored photo-realistic novel views. We are the first to propose a method to decompose neural radiance fields into multiple pure-colored layers for recoloring. The decomposed layer are jointly optimized with a learnable palette to produce more disjoint decomposition and more representative colors in the palette. The layers are then stacked with alpha blending to generate color radiance and render the final photo-realistic images. Recoloring a scene is as simple as altering the color in palettes.
\paragraph{Limitations} The RecolorNeRF is only designed for overall color editing rather than instance-level editing. Since layer decomposition is merely based on colors, two different objects with similar colors will be decomposed into one layer. In this case, their color cannot be changed separately.
% By introducing semantic segmentation, we can further decompose the scene into multiple layers by exploiting the semantics. This will be the future enhancement of RecolorNeRF.
Thus, one of the future improvements of RecolorNeRF is to incorporate semantics into layer decomposition.
%The RecolorNeRF is only suitable for color editing, not object or semantic editing. The color determines the layer decomposition entirely. As a result, two different objects of the same color will be decomposed into a single layer. Their color cannot be changed separately in this case. We can decompose the scene into multiple layers based on semantic information by introducing semantic segmentation; this will be a future enhancement of RecolorNeRF.

\begin{acks}
This work was supported in part by Shenzhen Portion of Shenzhen-Hong Kong Science and Technology Innovation Cooperation Zone under HZQB-KCZYB-20200089, in part by Hong Kong Research Grants Council Project No. 24209223, in part by Science, Technology and Innovation Commission of Shenzhen Municipality Project No. SGDX20220530111201008, in part by NSFC-62172348, in part by Outstanding Young Fund of Guangdong Province with No. 2023B1515020055, in part by Shenzhen General Project with No. JCYJ20220530143604010, and in part by CCF-Tencent Open Research Fund.
\end{acks}

\bibliographystyle{ACM-Reference-Format}
\balance
\bibliography{sample-base}

\appendix

\section{Implementation Details}

We implement our entire model and training pipeline with PyTorch \cite{paszke2019pytorch}. Our implementation is based on TensoRF \cite{chen2022tensorf} without any customized CUDA kernels.

\paragraph{TensoRF Backbone} We utilize TensoRF-VM as our backbone for querying density and opacity values, which offers a more compact scene representation compared to pure grid-based representations \cite{yu_and_fridovichkeil2021plenoxels,sun2022direct} and achieves faster training convergence than the original pure MLP-based representations \cite{mildenhall2021nerf,barron2021mip}. To disentangle the geometry and appearance of the scene, we construct separate density field $\mathcal{T}_\sigma$ and opacity field $\mathcal{T}_c$. Specifically, TensoRF-VM factorizes the geometry of the 3D space as a sum of outer products of vectors and matrices, as shown in Eq. \ref{eq:tensorfvm}.
% \begin{equation}
%     \mathcal{T}_\sigma = \sum_{r=1}^{R^\sigma_1} \boldsymbol{v}_{\sigma, r}^X \circ \boldsymbol{M}_{\sigma, r}^{Y,Z} + \sum_{r=1}^{R^\sigma_2} \boldsymbol{v}_{\sigma, r}^Y \circ \boldsymbol{M}_{\sigma, r}^{X,Z} + \sum_{r=1}^{R^\sigma_3} \boldsymbol{v}_{\sigma, r}^Z \circ \boldsymbol{M}_{\sigma, r}^{X,Y}
%     \label{eq:tensorfvm}
% \end{equation}
\begin{equation}
    \mathcal{T}_\sigma = \sum_{m\in[3]}\sum_{r=1}^{R^\sigma_m} \boldsymbol{v}_{\sigma, r}^m \circ \boldsymbol{M}_{\sigma, r}^m
    \label{eq:tensorfvm}
\end{equation}
where $\circ$ is the outer product operator, $R^\sigma_1, R^\sigma_2$ and $R^\sigma_3$ denote the number of components, $\boldsymbol{v}_{\sigma, r}^1, \boldsymbol{v}_{\sigma, r}^2, \boldsymbol{v}_{\sigma, r}^3\in \mathbb{R}^N$ are vectors along the $x$, $y$, and $z$ axes, and $\boldsymbol{M}_{\sigma, r}^1, \boldsymbol{M}_{\sigma, r}^2, \boldsymbol{M}_{\sigma, r}^3\in \mathbb{R}^{N\times N}$ represent matrices corresponding to the $y-z$, $x-z$, and $x-y$ planes, respectively.
While for the opacity field, TensoRF-VM factorizes a 4D feature tensor, incorporating an additional mode to represent the channel dimension, which boils down to Eq. \ref{eq:tensorfvm_app}:
% \begin{equation}
%     \mathcal{T}_c = \boldsymbol{B} \left ( \bigoplus_{r=1}^{R^c_1} \boldsymbol{v}_{c, r}^X \circ \boldsymbol{M}_{c, r}^{Y,Z} \oplus \bigoplus_{r=1}^{R^c_2} \boldsymbol{v}_{c, r}^Y \circ \boldsymbol{M}_{c, r}^{X,Z} \oplus \bigoplus_{r=1}^{R^c_3} \boldsymbol{v}_{c, r}^Z \circ \boldsymbol{M}_{c, r}^{X,Y}\right )
%     \label{eq:tensorfvm_app}
% \end{equation}
\begin{equation}
    \mathcal{T}_c = \boldsymbol{B} \left [ \bigoplus_{m\in[3]}\bigoplus_{r=1}^{R^c_m} \boldsymbol{v}_{c, r}^m \circ \boldsymbol{M}_{c, r}^m \right ]
    \label{eq:tensorfvm_app}
\end{equation}
where $\oplus$ is the concatenation operator, $\boldsymbol{v}_{c, r}^{m}$ and $\boldsymbol{M}_{c, r}^{m}$ are vector and matrix factors of the opacity field, and $\boldsymbol{B}$ is a feature basis matrix. After obtaining multi-channel features, we feed the feature into a shallow 3-layer MLP to yield the opacity for each item in the palette. In our experiment settings for $360^{\circ}$ synthetic scenes, we set $R^c_1 = R^c_2 = R^c_3 = 48$ for opacity fields and $R^\sigma_1 = R^\sigma_2 = R^\sigma_3 = 16$ for density fields. The resolutions of the vector and matrix factors $N$ are initially set to 128 and progressively increased to 300 over the training procedure. As for forward-facing scenes, we set $R^c_1=R^c_2=12, R^c_3=48$ for opacity fields and $R^\sigma_1=R^\sigma_2=4, R^\sigma_3=16$ for density fields. The vector and matrix factors are gradually upsampled from $128$ to $640$ over the training.

\paragraph{Optimization} Our training procedure follows the original TensoRF pipeline, enhanced by incorporating our proposed regularizers for palette learning and layer sparsity. Additionally, we apply a TV (total variation) loss and L1 norm penalty on the vector and matrix factors. The L1 norm weight is initially set to 8e-5 and reduced to 4e-5 after the first upsampling of the resolutions. The weight of the TV loss varies between ${1, 50}$ depending on scene complexity. For optimization, we utilize the Adam optimizer \cite{kingma2014adam} and train our models for 30,000 iterations. All experiments are conducted on a single NVIDIA RTX 3090 GPU.

\begin{figure*}[t]
	\centering
	\includegraphics[width=\textwidth]{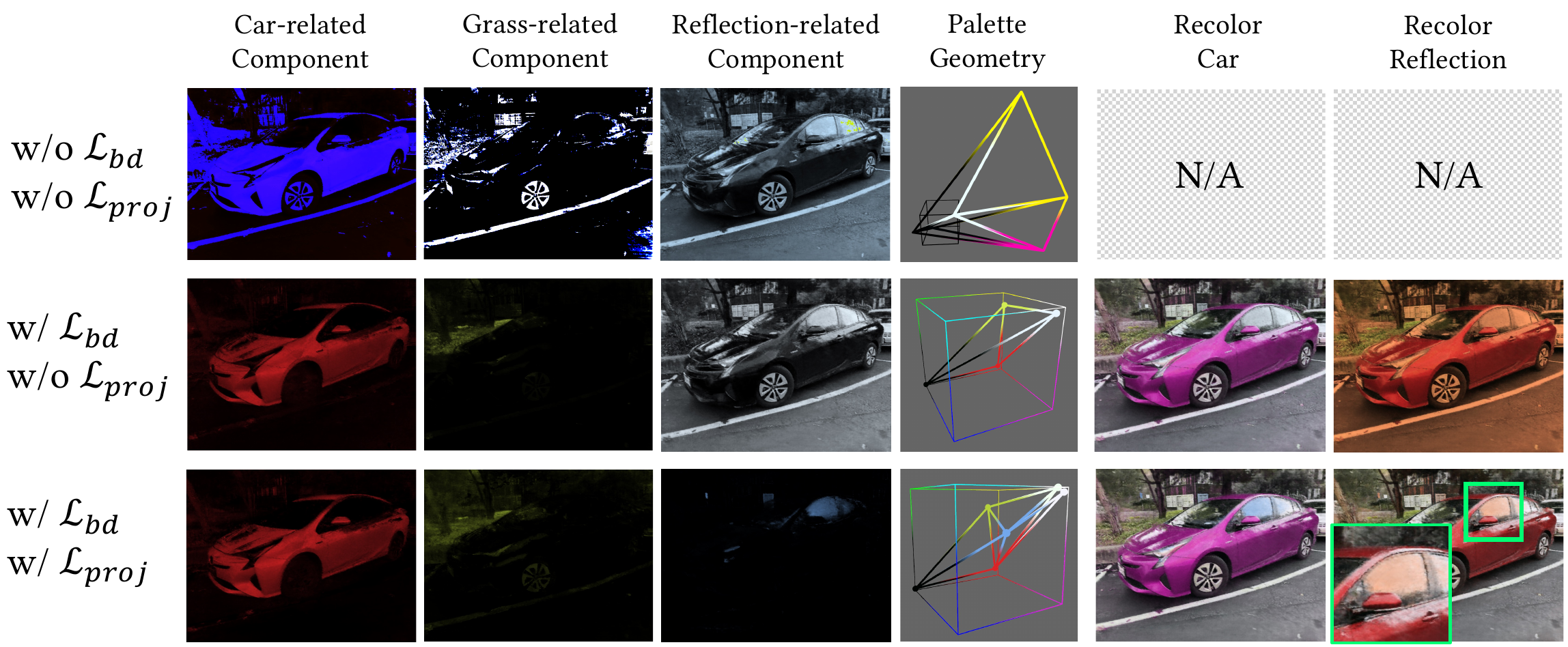}
	\vspace*{-18px}
	\caption{Ablation study on palette learning. In the first three columns, we visualize color components for three editing targets: car, grass, and reflection on the car window. The geometries of the optimized palettes are visualized in the fourth column, where the cubes represent the $[0,1]^3$ RGB space. The last two columns showcase color editing results.}
	\label{fig:res_ablation_plt}
	\vspace*{-8px}
\end{figure*}

\section{Baseline Implementations}

In all of our baseline implementations, we reuse their officially provided code and hyper-parameters.

\paragraph{NeRF + Image PCE}
In order to contrast our method with pure image-level recoloring, we employ a state-of-the-art palette-based image color editing method \cite{tan2016decomposing} as the baseline. To ensure fair comparisons, we synthesize novel views using the TensoRF backbone and then utilize the original implementation provided by the authors to generate recoloring results for the scheme NeRF + Image PCE. In this original algorithm, an independent palette is calculated for each novel view. In our enhanced scheme, NeRF + Image PCE with unified palette, we modify the code to gather pixels from all training views and generate a unified palette offline prior to processing each individual view. During the editing process, colors in the target palettes are matched with the nearest colors in the edited palettes of our method.
% We use the state-of-the-art palette-based image color editing method \cite{tan2016decomposing} as the baseline to contrast our method with pure image-level recoloring. To ensure a fair comparison, novel views are first synthesized using the TensoRF backbone. Then we utilize the authors' original implementation to generate recoloring results for the scheme NeRF + Image PCE. The original algorithm calculates an independent palette for each novel view. While in the enhanced scheme NeRF + Image PCE w/ unified palette, we modify the code to collect pixels from all training views and generate a unified palette offline before processing each individual view. During editing, colors in the target palettes are matched with the nearest colors in the edited palettes of our method.

\paragraph{CLIP-NeRF + DFF} The implementation of this baseline method is derived from \citep{kobayashi2022distilledfeaturefields}. Following their instructions, we extract semantic features from the training views and distill these semantics into neural fields. For the editing process, we prepare text prompts that describe our desired editing goals and optimize the scene by minimizing the distance between the embeddings of the text prompts and the embeddings of the rendered images. Additionally, we attempt to filter out rays during the optimization that do not intersect with the editing objects.

\paragraph{PosterNeRF} PosterNeRF \cite{tojo2022recolorable} is currently implemented only on synthetic NeRF datasets, leading us to exclude its results on LLFF, NVSF, and Tank and Temples datasets. To ensure fair comparisons, we transform the camera poses from the original datasets into the coordinate systems of the PosterNeRF renderer, enabling consistent viewpoints across the experiments. For recoloring PosterNeRF, we begin by substituting the closest color in the extracted palette of PosterNeRF with the new RGB value from RecolorNeRF's edited palette. Subsequently, we refine the palettes to achieve better results aligned with our desired editing goals.

\section{Experiments}

\subsection{Ablation Study}  We conduct an ablation study on the regularizers in our proposed palette learning. In the first row of Fig. \ref{fig:res_ablation_plt}, we disable the entire regularization on the palette learning. We can observe all the color components are over-saturated and the palette geometry is not included in the color space, i.e., unit cube $[0,1]^3$. Consequently, such a palette, intended for color editing, is deemed infeasible.
To address this issue, we introduced the convex hull bounding regularization $\mathcal{L}_{hull}$. This regularization ensures that the palette is contained within the color space. As shown in the second row of Fig. \ref{fig:res_ablation_plt}, the scene is successfully decomposed with the regularized palette, which leads to the success of recoloring the car. However, we find this regularizer tends to collapse the palette. In the exhibited case, the palette only consists of two editable colors, excluding the base color components in white and black, as dictated by the palette geometry. The sky-blue component of the reflection on the car window is absorbed into the white component, causing the recoloring of the reflection to fail. To overcome this issue, we further incorporate the projection-based convex regularization $\mathcal{L}_{proj}$, which forces colors in the palette to span a convex hull without overlapping vertices. With the addition of this regularizer, the sky-blue component is separated from the white component and forms a new vertex on the palette geometry, as shown in the last row of Fig. \ref{fig:res_ablation_plt}. This allows for a recoloring capability for changing the reflection color on the car window into ``sunset red''.

\subsection{Comparisons}

\begin{figure*}[p]
	\centering
	\includegraphics[trim={0 5 0 10},clip,width=0.95\textwidth]{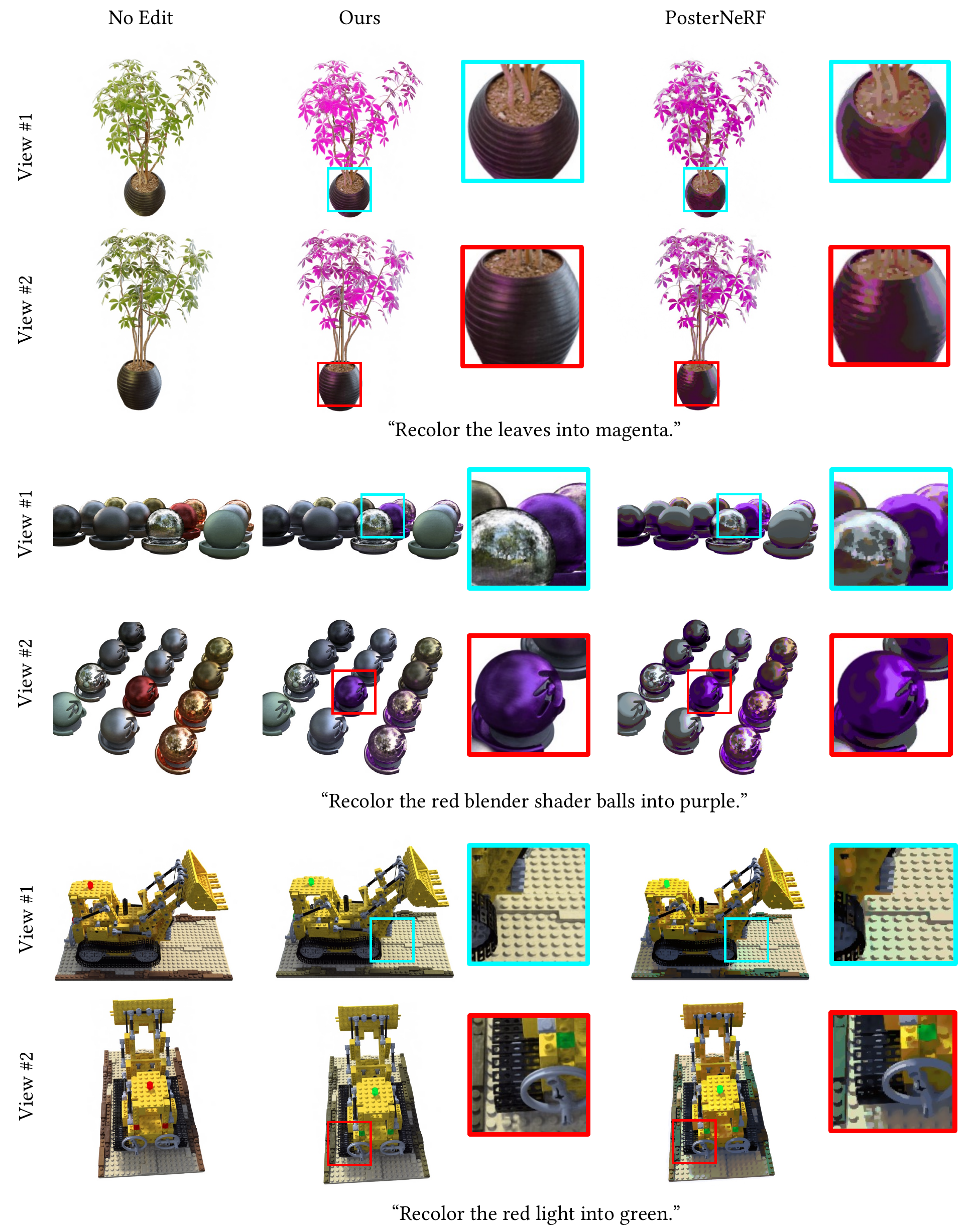}
	\vspace*{-8px}
	\caption{Comparisons of our method and PosterNeRF. Close-ups display the result details of each method. The text below each case is the desired editing goal.}
	\label{fig:res_cmp2}
\end{figure*}

\begin{figure*}[p]
	\centering
	\includegraphics[width=\textwidth]{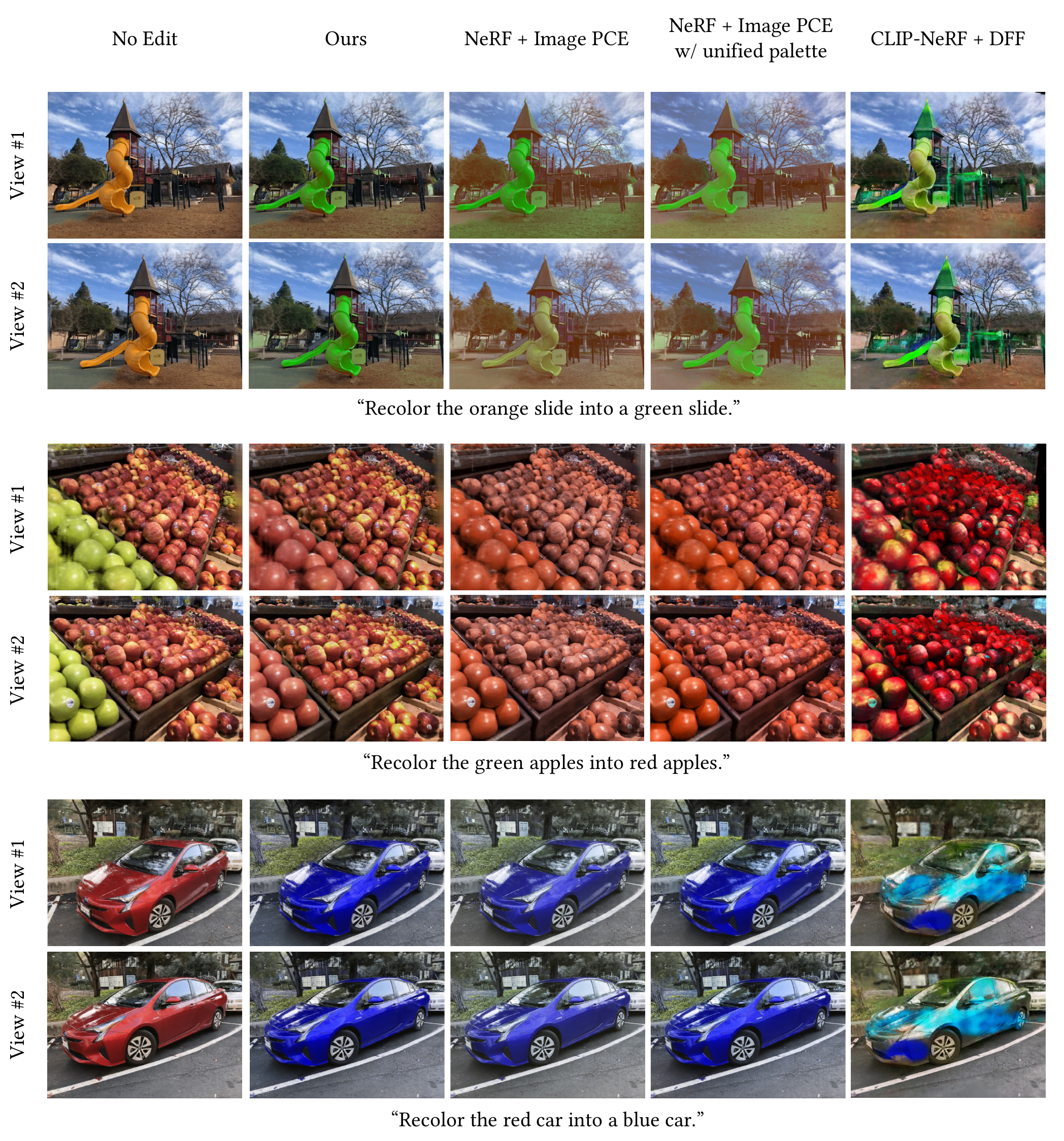}
	\caption{Comparisons of our method, NeRF + Image PCE, NeRF + Image PCE w/ unified palette, and CLIP-NeRF + DFF. The text below each case is the desired editing goal.}
	\label{fig:res_cmp}
\end{figure*}

\paragraph{Baselines} Fig. \ref{fig:res_cmp2} and \ref{fig:res_cmp} showcase more typical comparisons between our method and the baselines, which have been included in our user study for qualitative evaluation. As shown in the exhibited cases, our method outperforms the baselines on most of the complex scenes in terms of controllability and rendering quality.

\begin{figure*}[t]
	\centering
	\includegraphics[trim={0 0 0 0},clip,width=\textwidth]{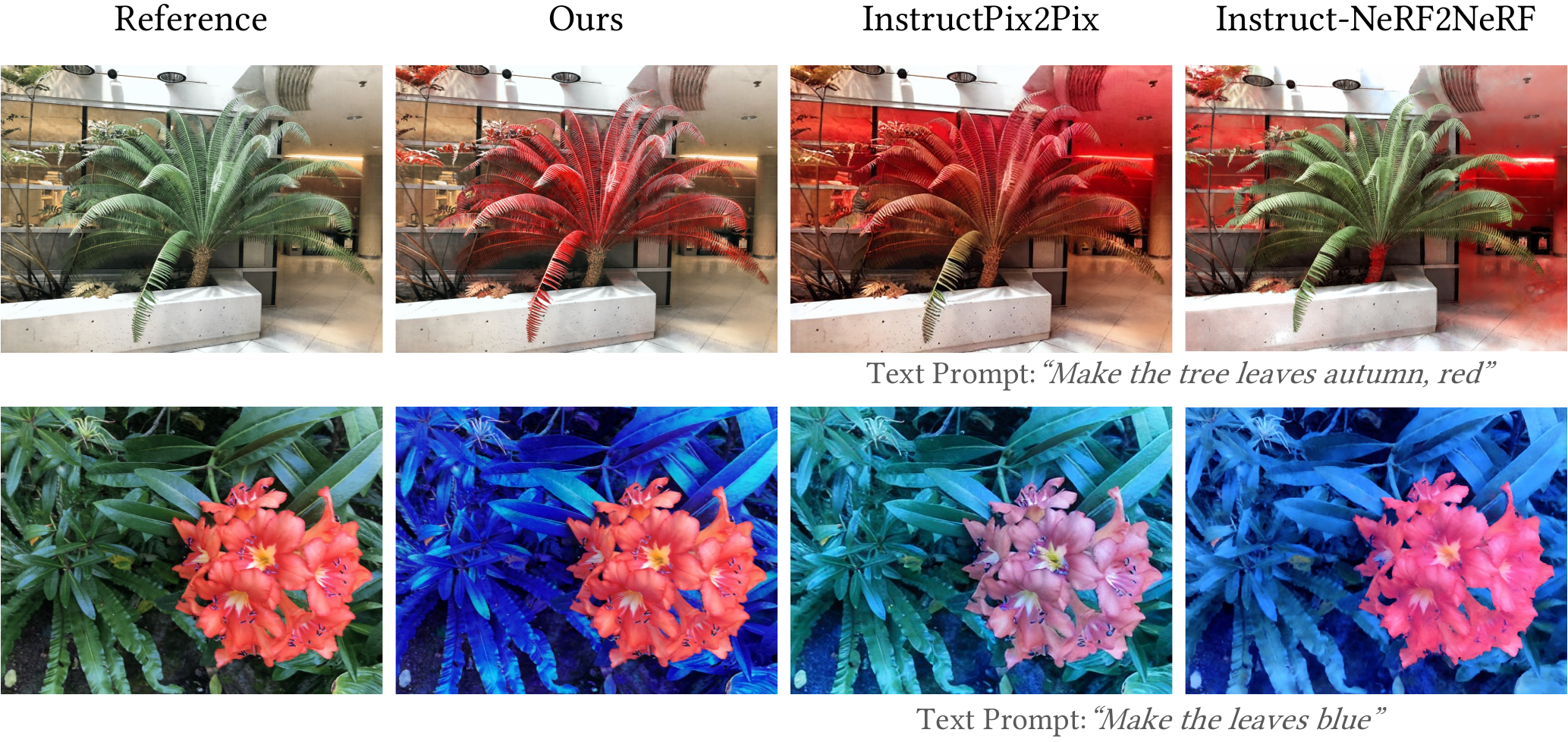}
	\vspace*{-18px}
	\caption{Comparisons of our method, InstructPix2Pix and Instruct-NeRF2NeRF. Text prompts for the two text-guided approaches are displayed below each case.}
	\label{fig:res_cmp_instruct}
	%\vspace*{-8px}
\end{figure*}

\paragraph{InstructPix2Pix and Instruct-NeRF2NeRF} To highlight the advantage of our proposed palette-based color editing method, we compare the latest diffusion-based editing approaches, InstructPix2Pix \cite{brooks2023instructpix2pix} and Instruct-NeRF2NeRF \cite{haque2023instruct}. These approaches enable image-level and NeRF-level color editing, respectively. Fig. \ref{fig:res_cmp_instruct} presents the comparisons of our method, InstructPix2Pix, and Instruct-NeRF2NeRF. Both of the two text-guided comparison methods suffer from severe ``color pollution'' problems, where areas out of the editing target are significantly affected. This is due to the failure of input text prompts to precisely guide the diffusion-based generation procedure toward the intended target. In contrast, our method, which leverages palettes, enables more precise and controllable color editing.

\begin{figure*}[t]
	\centering
	\includegraphics[trim={0 5 0 0},clip,width=\textwidth]{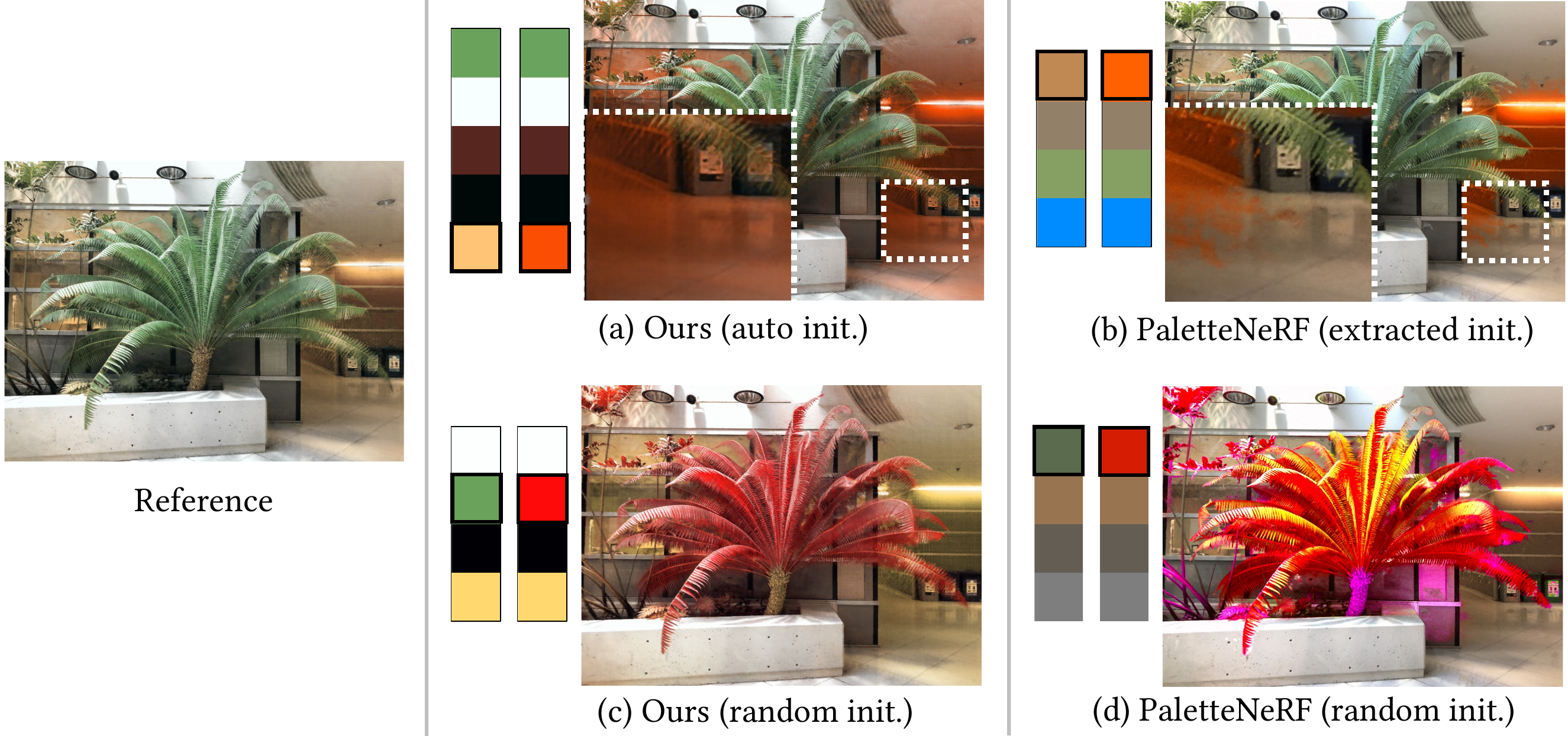}
	%\vspace*{-16px}
	\caption{Comparisons of our method and PaletteNeRF. We evaluate two initialization schemes of palettes: initialization from extracted palettes (Extracted Init.) and random initialization (Random Init.).}
	\label{fig:res_cmp_palettenerf}
	%\vspace*{-8px}
\end{figure*}

\paragraph{PaletteNeRF} We also compare our method with the concurrent work PaletteNeRF \cite{kuang2022palettenerf}.
% Fig. \ref{fig:res_cmp_palettenerf} shows the recoloring comparisons on the ``fern'' scene. First, we aim to recolor the background lighting into ``cadmium red''. We can observe PaletteNeRF generates some artifacts in the lighting and shadows (Fig. \ref{fig:res_cmp_palettenerf}b). While our method processes those transition areas of lighting and shadows in a smoother and more natural way (Fig. \ref{fig:res_cmp_palettenerf}a).
Fig. \ref{fig:res_cmp_palettenerf} presents the recoloring comparisons on the "fern" scene. Firstly, we aimed to recolor the background lighting to "cadmium red". It can be observed that PaletteNeRF introduces some artifacts in the lighting and shadows (Fig. \ref{fig:res_cmp_palettenerf}b). In contrast, our method handles these transitional areas of lighting and shadows more smoothly and naturally (Fig. \ref{fig:res_cmp_palettenerf}a).
% Second, we aim to highlight the effectiveness of our palette learning scheme even though palettes are randomly initialized. The editing goal is to recolor the fern into the red. We can observe our method can still yield considerable recoloring results (Fig. \ref{fig:res_cmp_palettenerf}c) despite the palette being randomly initialized. This is because our palettes can be fully learnable and obtained jointly with radiance field optimization. However, PaletteNeRF requires a time-consuming two-stage training procedure, where an additional vanilla NeRF should be trained in advance, in order to extract initial palettes, similar to \cite{tojo2022recolorable}. Without the pre-computed palette, the color editing results of PaletteNeRF degenerate (Fig. \ref{fig:res_cmp_palettenerf}d).
Secondly, we aimed to validate the effectiveness of our palette learning scheme, even when the palettes are randomly initialized. Our goal is to recolor the fern to red.
It is worth noting that PaletteNeRF requires a time-consuming two-stage training procedure. Prior to training a PaletteNeRF, an additional vanilla NeRF model is trained to extract initial palettes, similar to the approach described in \cite{tojo2022recolorable}. Without the pre-computed initial palette, the color editing results of PaletteNeRF degrade significantly (Fig. \ref{fig:res_cmp_palettenerf}d)), revealing its deficiency in palette optimization. However, our method still achieves considerable recoloring results (Fig. \ref{fig:res_cmp_palettenerf}c) despite the random initialization of the palette. This is attributed to our fully learnable palettes, which are obtained jointly with radiance field optimization.

\subsection{More Visual Results}

\begin{figure*}[p]
	\centering
	\includegraphics[width=\textwidth]{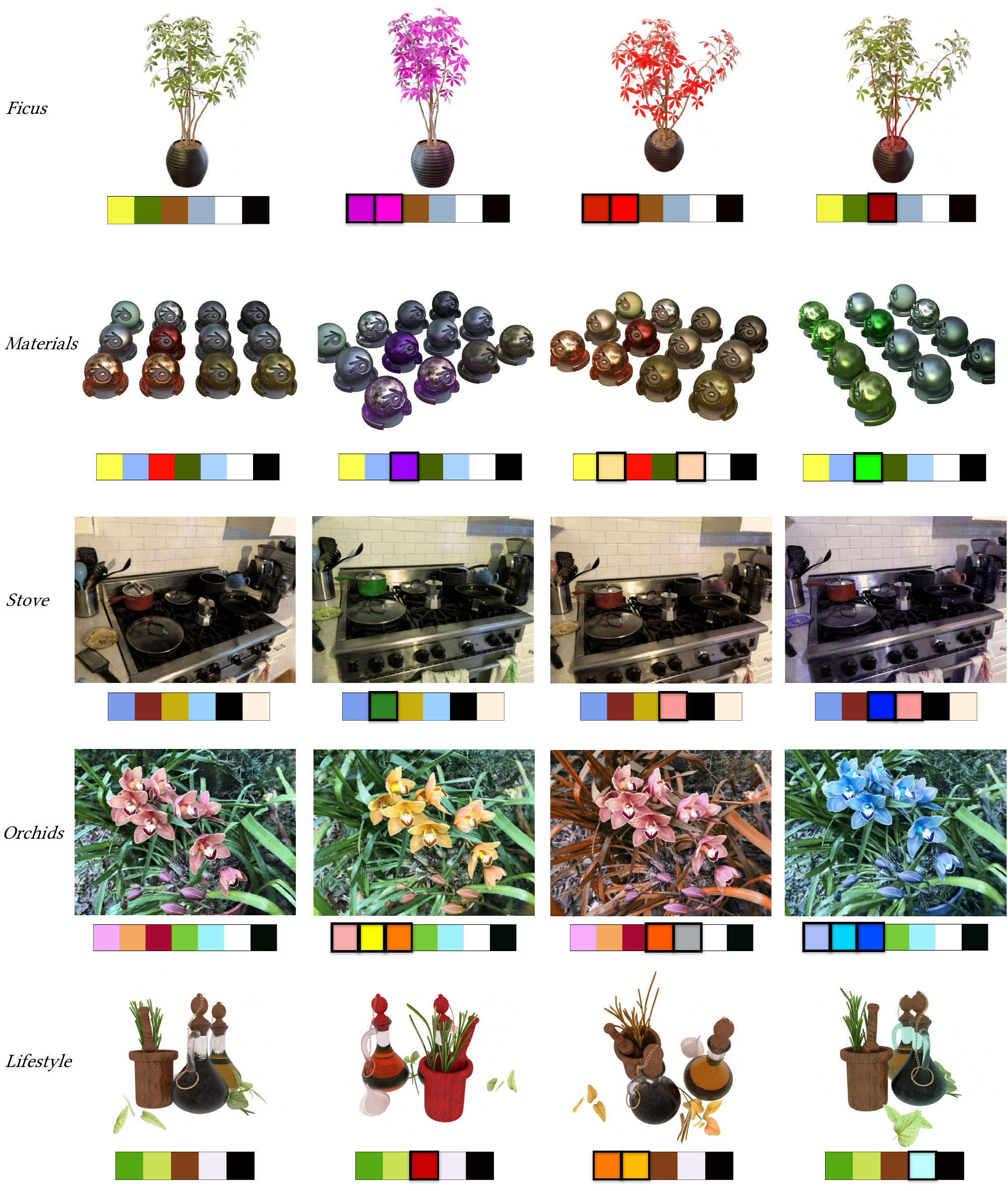}
	\caption{Gallery of our color editing results. The first image of each row is the reference before editing along with the optimized palette beneath. The other 3 images showcase 3 examples of color editing with the corresponding edited palettes.}
	\label{fig:res_vis2_2}
\end{figure*}

\begin{figure*}[p]
	\centering
	\includegraphics[width=\textwidth]{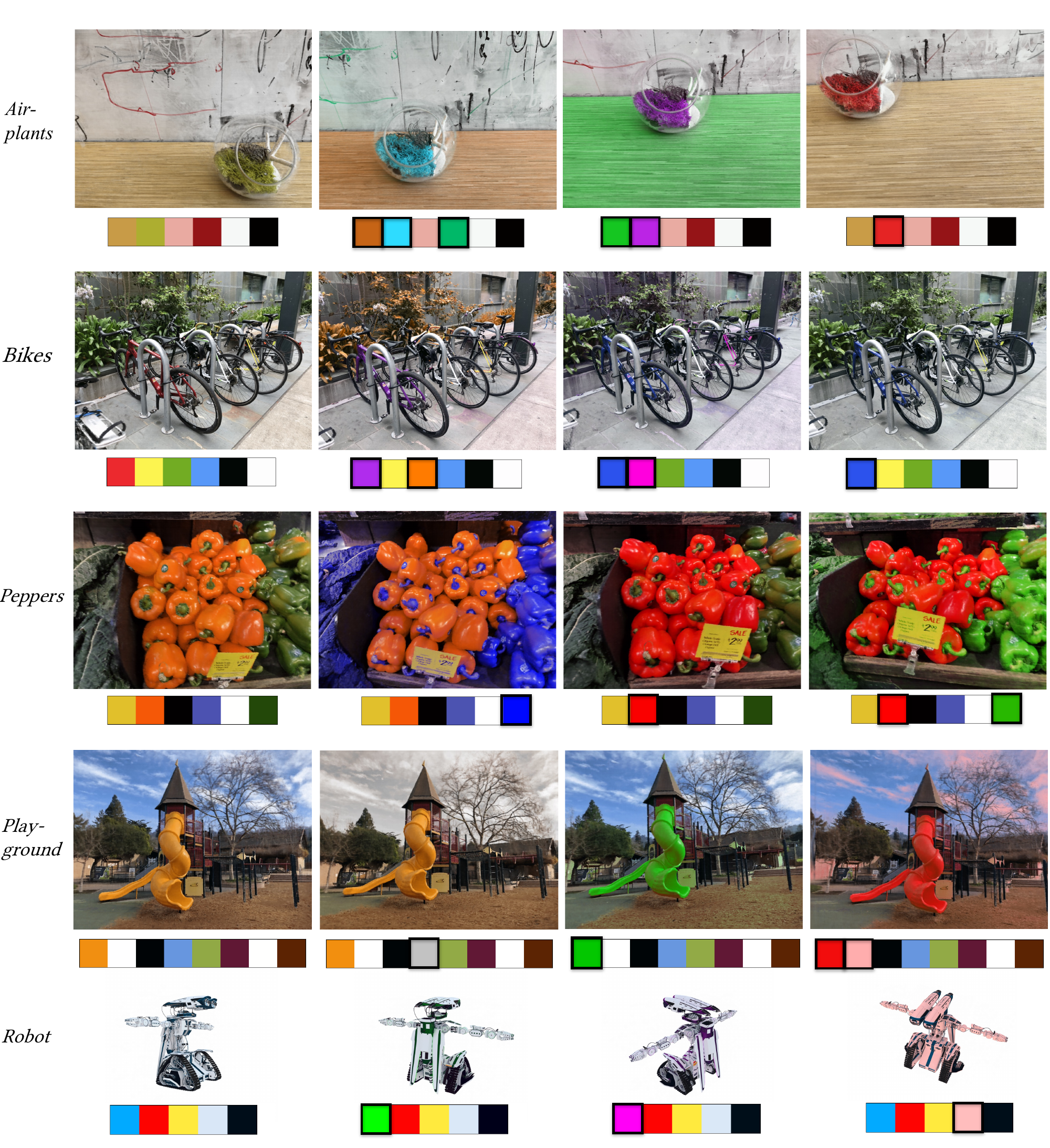}
	\caption{(cont.) Gallery of our color editing results. The first image of each row is the reference before editing along with the optimized palette beneath. The other 3 images showcase 3 examples of color editing with the corresponding edited palettes.}
	\label{fig:res_vis2}
\end{figure*}

We present additional visual results in Fig. \ref{fig:res_vis2_2} and \ref{fig:res_vis2} to validate the effectiveness of our method, while more illustrative examples can be found in our supplementary demo videos.

\end{document}